\documentclass[dvipsnames]{article} %
\PassOptionsToPackage{numbers,sort&compress}{natbib}
\usepackage{colm2024_conference}
\setcitestyle{numbers,square,citesep={,}}
\makeatletter
\AtBeginDocument{\let\NAT@space\relax}
\makeatother

\usepackage{booktabs}
\usepackage{enumitem}
\usepackage{wrapfig}
\usepackage{placeins} %
\usepackage{needspace} %
\usepackage{algorithm}
\usepackage{algpseudocode}
\usepackage{graphicx}
\usepackage{multicol}

\usepackage{microtype}
\usepackage{amsmath}
\usepackage{amssymb}   %
\usepackage{colortbl}
\usepackage[utf8]{inputenc}
\usepackage[T1]{fontenc}
\definecolor{lightgray}{rgb}{0.9,0.9,0.9}
\usepackage{caption}
\usepackage{subcaption}
\usepackage{setspace}
\usepackage{url}
\usepackage{multirow}
\usepackage{tabularx}
\usepackage{blindtext}
\usepackage{pgfplots}
\pgfplotsset{compat=1.18}
\usepackage{tikz}
\usetikzlibrary{er,positioning,bayesnet}
\usetikzlibrary{arrows.meta,fit,backgrounds,calc}   %
\usepackage{makecell}
\usepackage[safe]{tipa}
\usepackage{siunitx}
\usepackage{nicefrac}
\usepackage{listings}
\usepackage[raster,skins, most]{tcolorbox} %
\usepackage{xltabular}
\usepackage{adjustbox}
\usepackage{xurl}
\usepackage{rotating}
\usepackage[normalem]{ulem}
\usepackage{caption}
\usepackage{subcaption}
\usepackage[subfigure]{tocloft}
\usepackage{fontawesome}
\usepackage{arydshln}
\usepackage{pifont}

\useunder{\uline}{\ul}{}

\renewcommand*\ttdefault{cmtt}
\providecommand{\ghlink}{}

\renewcommand{\ghlink}{https://github.com/Tongyi-MAI/MAI-Gym}

\newcommand*\justify{%
  \fontdimen2\font=0.4em%
  \fontdimen3\font=0.2em%
  \fontdimen4\font=0.1em%
  \fontdimen7\font=0.1em%
  \hyphenchar\font=`\-%
}

\renewcommand{\texttt}[1]{%
  \begingroup
  \ttfamily
  \begingroup\lccode`~=`/\lowercase{\endgroup\def~}{/\discretionary{}{}{}}%
  \begingroup\lccode`~=`[\lowercase{\endgroup\def~}{[\discretionary{}{}{}}%
  \begingroup\lccode`~=`.\lowercase{\endgroup\def~}{.\discretionary{}{}{}}%
  \catcode`/=\active\catcode`[=\active\catcode`.=\active
  \justify\scantokens{#1\noexpand}%
  \endgroup
}

\usepackage{makecell}

\definecolor{lightblue}{RGB}{173,216,230}
\definecolor{uclablue}{RGB}{159, 195, 224}

\definecolor{uclagold}{RGB}{255, 240, 180}

\definecolor{aliceblue}{RGB}{255, 238, 241}

\definecolor{cadmiumgreen}{rgb}{0.0, 0.42, 0.24}

\definecolor{myred}{rgb}{0.7, 0.3, 0.0}
\definecolor{myblue}{rgb}{0.2, 0.3, 0.6}
\definecolor{babygreen}{rgb}{0.85, 0.97, 0.85}

\definecolor{purple1}{RGB}{126, 107, 196}
\definecolor{purple2}{RGB}{199, 158, 207}
\definecolor{purple3}{RGB}{214, 200, 255}
\definecolor{purple4}{RGB}{254, 240, 255}

\definecolor{deepblue}{RGB}{48, 58, 82}
\definecolor{light_purple}{RGB}{228, 228, 251}
\definecolor{deep_purple}{RGB}{234, 224, 240}
\definecolor{deep_purple2}{RGB}{102, 48, 226}
\definecolor{mairow}{HTML}{CFCDFD} %
\makeatletter
\newcommand{\fullrowcolor}[1]{\noalign{{\color{#1}\hrule\@height\dimexpr\ht\@arstrutbox+\dp\@arstrutbox\relax}\vskip-\dimexpr\ht\@arstrutbox+\dp\@arstrutbox\relax}}
\makeatother

\definecolor{deepPurple}{HTML}{330066}

\definecolor{uclablue_old}{rgb}{0.15, 0.45, 0.68}
\hypersetup{
    breaklinks,
    citecolor=uclablue_old,
    colorlinks=true,
}
\usepackage[most]{tcolorbox}
\newtcbtheorem{promptbox}{Prompt}{
    breakable,
    colframe=deepPurple!70,
}{box}

\newtcolorbox{mybox}[2][]
  {colback = black!5!white, colframe = black!75!black, fonttitle = \bfseries,
    colbacktitle = black!100!black, enhanced, before upper={\fontsize{8}{11}\obeyspaces\obeylines\selectfont}, fontupper=\selectfont,
    attach boxed title to top left={yshift=-2.2mm,xshift=4mm},
    title=#2,#1}

\IfFileExists{logo/tongyi.jpg}{%
}{%
}

\providecommand{\sys}{MAI-Gym}

\providecommand{\mobileworld}{MobileWorld}

\newcommand{\modelname}{MAI-Gym}
\title{
    \modelname{}: An Open All-in-One Platform for GUI Agent Evaluation and RL Training across Mobile and Computer Environments
    \vspace{-0.2em}
}

\author{%
  {\fontsize{10pt}{11pt}\selectfont
Alibaba Token Hub, Alibaba Group}\thanks{MAI-UI Team. MAI-GYM was used to train Qwen-UI-Agent~\citep{qwenuiagent}. See \hyperref[contribution]{Contributions} for the full author list.}
}

\makeatletter
\def\addcontentsline#1#2#3{%
  \addtocontents{#1}{\protect\contentsline{#2}{#3}{\thepage}{\@currentHref}\protected@file@percent}}
\makeatother

\begin{document}

\maketitle

\begin{abstract}
\vspace{-0.5em}

Online reinforcement learning (RL) with verifiable rewards has driven
rapid progress in mathematics, coding, and tool use, but scaling
this success to GUI agents remains challenging due to the high cost
of interactive environments and the scarcity of realistic training
tasks. We introduce \textbf{\sys{}}, an open all-in-one platform
that unifies GUI agent evaluation, data collection, and online RL
training across mobile, computer, and web environments.
Through a shared rollout pipeline, \sys{} ensures consistent agent
behavior across evaluation and training, supports ten established
benchmarks, and scales to $20{,}000$ parallel environment instances.
Beyond scalable infrastructure, \sys{} enables automated synthesis
of diverse GUI applications and challenging tasks, including
resettable mocked apps and coherent cross-app worlds with
execution-validated task verifiers.
Using \sys{}, we train \textbf{MAI-Gym-9B}, a unified 9B agent
for mobile and computer use, achieving \textbf{state-of-the-art
performance among open GUI agents under 10B parameters}:
$78.4\%$ on \mobileworld{} and $76.2\%$ on OSWorld-Verified.
Importantly, RL in a mocked train-ticketing app transfers to
its real-world counterpart, improving task success on a physical
phone from $28.8\%$ to $48.1\%$.
We release the framework, synthesis skills, mocked apps,
$2{,}000$ verified tasks, and the model checkpoint to support reproducible research and further advances in scalable GUI agent training.

\end{abstract}

\begin{figure}[H]
    \centering
    \includegraphics[width=0.99\linewidth]{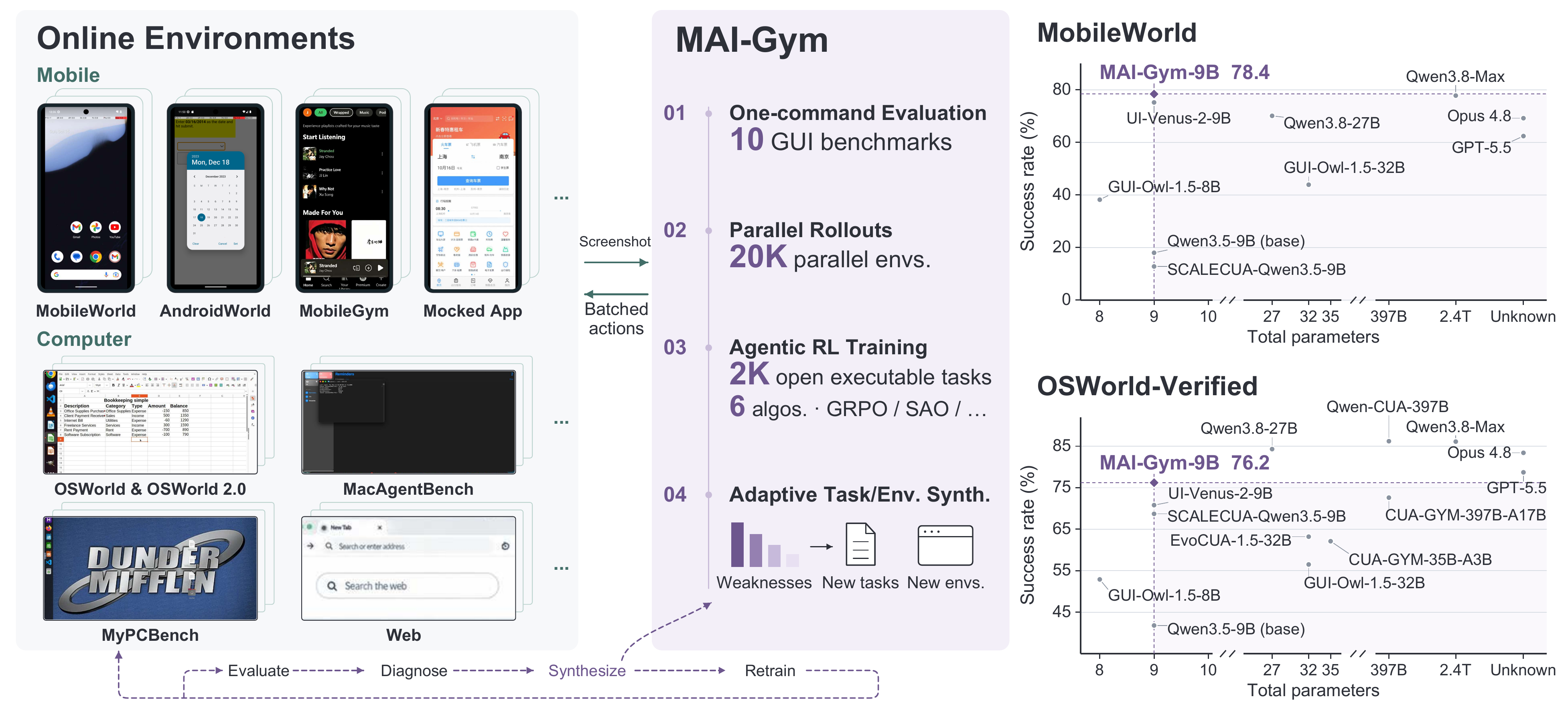}
    \caption{\textbf{Overview of \sys{}.} Left: online GUI environments. Center: core capabilities. Bottom: the evaluation--diagnosis--synthesis--retraining loop. Right: performance comparison of MAI-Gym-9B with models of different sizes on \mobileworld{} and OSWorld-Verified.}
    \label{fig:teaser}
\end{figure}

\clearpage
\tableofcontents
\clearpage

\section{Introduction}
\label{sec:intro}

Reinforcement learning (RL) with verifiable rewards has driven rapid
progress in mathematics, coding, and tool use~\citep{deepseekr1,swegym,
r2egym,swesmith,envscaler,agentworld}. Beyond advances in training
algorithms, this progress relies on environments that are inexpensive
to execute, easy to verify, and rich in meaningful tasks. Mathematical
problems often admit automatic answer checking, while software
repositories provide executable tasks with files and tests. Such
environments can efficiently support large-scale training and evaluation,
as many tasks can share the same verification or execution infrastructure.
Recent work has extended online RL to GUI agents for mobile and computer
use~\citep{digirl,mobilerl,webrl,computerrl,uitars2}. However,
\textbf{scaling GUI-based RL to a comparable level remains challenging
due to the unique demands of interactive GUI environments.}

Bringing this success to GUI agents presents two fundamental challenges.
\textbf{First, GUI environments are expensive to scale and difficult
to maintain consistently.}
Unlike a code sandbox, a GUI instance cannot be shared by concurrent attempts, because the agent acts on the whole screen and changes the device's files and settings. Each attempt therefore needs its own phone, computer, or browser instance in a known initial state, and makes many screenshot-based model calls before the instance is reset. Large-scale training runs thousands of such instances at once. Because platforms and benchmarks differ, evaluation and training easily end up with separate agent implementations that drift apart, so measured performance can depend on the harness as well as the model~\citep{structagent}. Open infrastructure is therefore needed that runs environments at scale and keeps agent behavior the same in evaluation and training.
\textbf{Second, realistic GUI learning requires diverse tasks grounded in coherent and resettable environment states.} Unlike isolated question-answering tasks, real-world GUI workflows often depend on information distributed across multiple applications and files. For example, replying to an email with the total from an attached invoice requires the relevant email, attachment, and invoice contents to exist consistently in the environment. However, existing task synthesis approaches often rely
on isolated input files or shared sample data (Table~\ref{tab:gym_overview}), limiting the diversity and realism of generated tasks. Furthermore, real-world applications involving transactions can be unsafe or impractical for training, while evaluating adaptation to unfamiliar interfaces requires entirely new applications. Addressing these challenges calls for both scalable execution infrastructure and automated construction of realistic, resettable GUI environments.

Existing GUI gyms typically address only part of this problem
(Table~\ref{tab:gym_overview}). Infrastructure-focused gyms make evaluation or training easy to reproduce~\citep{browsergym,webgym,mobilegym,scalecua_rl,clawgui}, but each covers only one platform, and those that release training code run at most hundreds of environments in parallel. In contrast, data-oriented gyms~\citep{cuagym,gymanything,guigenesis,phoneworld} curate many new apps and tasks, but release neither training code nor a way to run existing benchmarks. 
These limitations are closely intertwined: scalable infrastructure requires a continuous supply of meaningful tasks, while synthesized tasks are most useful when they can be efficiently evaluated and used for learning. \textbf{A unified platform that combines scalable execution with automated environment and task synthesis is therefore essential for advancing GUI agent RL.}

To this end, we introduce \textbf{\sys{}}, an open all-in-one platform for GUI agent evaluation and online RL training across mobile, computer, and web environments (Figure~\ref{fig:teaser}). Its central design choice is a shared rollout path: for a given agent and environment, evaluation, data collection, and training reuse the same observation processing, context construction, and action execution, so a fix to the agent code reaches all three at once. With a single command, researchers can evaluate a local or API-served model on ten existing benchmarks (three mobile, four computer, and three web) and inspect every trajectory in a dashboard. With another, they can launch online RL training, including joint training on mobile and computer tasks. 
The platform automatically manages environment provisioning, isolation, resetting, and recovery, and we have run it with $20{,}000$ environment instances in parallel.

\textbf{Beyond scalable infrastructure, \sys{} enables automated
synthesis of diverse applications and realistic, challenging tasks, expanding GUI learning beyond fixed benchmarks and environments.}
At the app level, coding agents construct \emph{mocked apps}: self-contained applications with resettable backends. They come in two kinds: \emph{real-world clones} of existing services, enabling safe practice of transactional workflows, and \emph{novel apps} with entirely new interfaces for studying agent adaptation.
At the task level, \sys{} introduces \textbf{cross-app world construction} within  \mobileworld{} and OSWorld environments. Rather than generating isolated tasks, it first constructs a coherent world from a persona and scenario, populating multiple applications with interconnected files and records that share consistent people, events, and facts. Tasks are then generated from this world, requiring agents to retrieve, reason over, and integrate information across applications.
To ensure reliable evaluation, task requirements are established
before generating code-based verifiers, which are subsequently
validated through execution on initial, completed, and reset states.
Furthermore, failure-driven synthesis uses diagnoses of unsuccessful agent trajectories to target model weaknesses. All synthesized applications and tasks share the same execution interface as existing benchmarks, \textbf{forming a unified workflow for environment synthesis, evaluation, and online RL training.}

\begin{table}[!t]
\centering
\captionsetup{font=small,skip=4pt}
\caption{\textbf{Comparison with representative online GUI gyms.}}
\label{tab:gym_overview}
\begingroup
\scriptsize
\setlength{\tabcolsep}{2pt}
\renewcommand{\arraystretch}{1.1}
\newcommand{\tcheck}{\ensuremath{\checkmark}}
\newcommand{\tx}{\ensuremath{\times}}
\newcommand{\tpart}{{\tiny\ding{119}}}
\newcommand{\poff}[1]{\textcolor{black!18}{#1}}
\newcommand{\plat}[3]{%
  \ifcase#1 \poff{\faMobile}\or\faMobile\or\faMobile\rlap{$^{s}$}\fi\hspace{0.45em}%
  \ifnum#2=1 \faDesktop\else\poff{\faDesktop}\fi\hspace{0.2em}%
  \ifnum#3=1 \faGlobe\else\poff{\faGlobe}\fi}
\newcommand{\bench}[2]{#1\,(#2)}

\setlength{\tabcolsep}{2pt}
\renewcommand{\arraystretch}{1.15}

\renewcommand{\tabularxcolumn}[1]{m{#1}}
\begin{tabularx}{\linewidth}{
  >{\raggedright\arraybackslash}X
  >{\centering\arraybackslash}m{0.077\linewidth}
  >{\centering\arraybackslash}m{0.114\linewidth}
  >{\centering\arraybackslash}m{0.081\linewidth}
  >{\centering\arraybackslash}m{0.169\linewidth}
  >{\centering\arraybackslash}m{0.23\linewidth}
  >{\centering\arraybackslash}m{0.103\linewidth}
}
\toprule
\multirow{2}{*}{\textbf{System}} &
\textbf{Scope} &
\multicolumn{2}{c}{\textbf{Open-source Infrastructure}} &
\multicolumn{3}{c}{\textbf{Tasks \& Environments}} \\
\cmidrule(lr){2-2}\cmidrule(lr){3-4}\cmidrule(l){5-7}
& Platforms & \shortstack{Benchmark\\evaluation} & \shortstack{Online RL\\(\# envs)}
& \shortstack{New\\environments} & \shortstack{New World \\ (files \& app state)} & Reward \\
\midrule
BrowserGym~\citep{browsergym}     & \plat{0}{0}{1} & \bench{$8$}{\faGlobe} & \tx & \tx & \tx & Code \\
WebGym~\citep{webgym}             & \plat{0}{0}{1} & \tx & \tcheck\,$256$ & \tx & \tx & VLM \\
MobileGym~\citep{mobilegym}       & \plat{2}{0}{0} & \tx & \tcheck\,$256$ & Mocked apps & Pre-defined World & Code \\
\mbox{GUI-Genesis}~\citep{guigenesis} & \plat{2}{0}{0} & \tx & \tx$^{\dagger}$ & Mocked apps & Generated per task, in app code & Code \\
PhoneWorld~\citep{phoneworld}     & \plat{1}{0}{0} & \tx & \tx$^{\dagger}$ & Mocked apps & Pre-defined World & Code \\
CUA-Gym~\citep{cuagym}            & \plat{0}{1}{1} & \tx & \tx$^{\dagger}$ & Mocked websites & Generated per task, lightweight & Code \\
\mbox{Gym-Anything}~\citep{gymanything} & \plat{1}{1}{1} & \tx & \tx & Real software & Pre-defined World & VLM \\
SCALECUA~\citep{scalecua_rl}      & \plat{0}{1}{0} & \bench{$2$}{\faDesktop} & \tcheck\,$600$ & \tx & \tx & Code \\
\midrule
\rowcolor{mairow}
\textbf{MAI-Gym} & \plat{1}{1}{1} & \bench{$\mathbf{10}$}{3\,\faMobile\;4\,\faDesktop\;3\,\faGlobe}
  & \tcheck\,$\mathbf{20{,}000}$
  & \textbf{Mocked apps \& websites}
  & \makecell[c]{\textbf{Generated cross-app worlds, }\\\textbf{shared by many tasks}} & \textbf{Code \& VLM} \\
\bottomrule
\end{tabularx}

\vspace{2pt}
\begin{minipage}{\linewidth}
\scriptsize
$^{s}$Simulated in a browser. $^{\dagger}$RL was run; code not released.
\end{minipage}
\endgroup
\vspace{-1.6em}
\end{table}

Using \sys{}, we train \textbf{MAI-Gym-9B}, a unified 9B policy for mobile and computer use, in one joint online RL run on synthesized tasks, and evaluate it on seven benchmarks. On \mobileworld{} and OSWorld-Verified, whose tasks never appear in training, it achieves state-of-the-art results among open GUI agents of comparable size. 
Moreover, RL consistently improves performance over the SFT warm-up checkpoint across all five benchmarks outside its training environments, demonstrating generalization beyond them.
We further investigate the \textbf{sim-to-real transferability} of capabilities learned in synthesized applications. Notably, training in real-world clones transfers effectively to the real apps: RL in the clone of a train-ticketing app improves success on $52$ tasks in the real app, evaluated on a physical phone, from $28.8\%$ to $48.1\%$. Training on entirely novel apps also improves held-out task success from $25.8\%$ to $45.2\%$,  highlighting the potential of synthesized environments for learning generalizable GUI capabilities. To sum up, our contributions are fourfold:

\begin{itemize}[leftmargin=*,topsep=0.4em,itemsep=0.35em,parsep=0pt]
    \item \textbf{A Unified and Scalable Platform for GUI Agent
Evaluation and Online RL.}
We introduce MAI-Gym, an open all-in-one platform that unifies
evaluation, data collection, and online RL training across mobile,
computer, and web through shared environment interfaces and agent
harnesses. It supports ten benchmarks and demonstrates
scalability to $20{,}000$ parallel environments
(Section~\ref{sec:maigym}).

    \item \textbf{Automated Application and Task Synthesis for Scalable GUI Learning.} 
    We introduce automated synthesis of resettable mocked applications and challenging cross-app tasks grounded in coherent environment states. Our synthesis pipeline constructs interconnected application data, generates execution-validated task verifiers, and supports failure-driven task generation (Section~\ref{sec:taskenv}).

    \item \textbf{The Strongest Open-Source GUI Agent under 10B Parameters.} We release MAI-Gym-9B, a unified mobile and computer agent achieving $78.4\%$ on \mobileworld{} and $76.2\%$ on OSWorld-Verified, setting new state-of-the-art results among open GUI agents under 10B parameters. It outperforms strong open baselines, including \mbox{UI-Venus-2-9B}~\citep{uivenus2} ($75.2$/$70.8$), \mbox{SCALECUA-Qwen3.5-9B}~\citep{scalecua_rl} ($12.8$/$68.7$), and even the larger \mbox{GUI-Owl-1.5-32B}~\citep{mobileagent35} ($43.9$/$56.5$) (Section~\ref{sec:exp}).

    \item \textbf{Open Resources for Reproducible GUI Research.} We release the framework, including evaluation harnesses and online-RL training code; synthesis skills; $49$ mocked apps with accompanying tasks; and $2{,}000$ tasks with execution-tested verifiers, $1{,}000$ each for \mobileworld{} and OSWorld.
\end{itemize}

\section{Preliminaries}
\label{sec:prelim}

This section defines the executable tasks that \sys{} runs (Section~\ref{sec:prelim:task}), the agent harness and its GUI and CLI action space (Section~\ref{sec:prelim:harness}), and online RL over these tasks (Section~\ref{sec:prelim:online}).

\subsection{Executable Tasks in GUI Agents}
\label{sec:prelim:task}

An \emph{environment} $\mathcal{E}$ is a software system in which an agent acts, such as an Android device, a desktop virtual machine, or a browser application. An \emph{executable task} is
\begin{equation}
\label{eq:task}
    x=(\mathcal{E},\, s_0,\, g,\, V),
\end{equation}
with an initial state $s_0$, a natural-language instruction $g$, and a verifier $V$. The initial state holds the application records, files, accounts, and settings, such as the current date, that are loaded before the episode begins. It is part of the task: the same instruction under a different $s_0$ can have a different correct outcome. Each attempt runs in its own isolated \emph{instance} of $\mathcal{E}$, reset to $s_0$.

\paragraph{Verification.}
The verifier checks the final state $s_T$ and, when needed, the trajectory $\tau$. We express a task's requirements as binary criteria $b_j(s_T,\tau)\in\{0,1\}$ with weights $w_j>0$, of which a subset $\mathcal{C}$ is required. The reward and task success are
\begin{equation}
\label{eq:reward}
    r=\frac{\sum_j w_j b_j}{\sum_j w_j}\in[0,1],
    \qquad
    S=\prod_{j\in\mathcal{C}} b_j ,
\end{equation}
so the reward gives partial credit while success requires every required criterion. Criteria can be checked by code that reads application or system state, or by a VLM judge; \sys{} supports both.

\subsection{Agent Harness and Action Space}
\label{sec:prelim:harness}

An agent is a policy together with its \emph{harness}, the code between the policy and the environment. At step $t$, the harness builds the context $c_t$ from the instruction, the interaction history, and the current observation $o_t$, a screenshot together with any command output. It then invokes the policy, parses the output $y_t$ into actions, and executes them. Changing the harness alone can change a fixed model's benchmark score~\citep{structagent}, which motivates the shared harness of Section~\ref{sec:arch}.

\begin{table}[!b]
\vspace{-0.4em}
\centering
\captionsetup{font=small,skip=4pt}
\caption{\textbf{Agent observations, context, and actions.}}
\label{tab:action_space}
\footnotesize
\setlength{\tabcolsep}{2pt}
\renewcommand{\arraystretch}{1.15}
\begin{tabular*}{\linewidth}{@{\extracolsep{\fill}}lccc@{}}
\toprule
\textbf{Agent} & \textbf{Observation}\,{\color{gray}$o_t$} & \textbf{Context}\,{\color{gray}$c_t$} & \textbf{Actions}\,{\color{gray}$y_t$} \\
\midrule
GUI agent & Screenshot & Last $k$ screenshots + text history & Click, type, drag, \ldots \\
CLI agent & Command output & Text history, often summarized & Shell commands \\
\midrule
\fullrowcolor{mairow}
\textbf{\sys{}} & \textbf{Screenshot + command output} & \textbf{Customized by the harness} & \textbf{Batched GUI + CLI actions} \\
\bottomrule
\end{tabular*}
\end{table}

\paragraph{GUI and CLI Actions.}
\sys{} agents act through both the GUI and the command line (Table~\ref{tab:action_space}). GUI actions follow Qwen-UI-Agent~\citep{qwenuiagent} (Appendix~\ref{app:actions}) and work with any application on screen. On computers, agents can also run shell commands and choose the interface for each action: one command can locate, edit, or move files that would take many GUI steps, while the GUI remains necessary for applications without a command-line interface. Mobile and web environments use GUI actions only. A failed command returns its error as feedback rather than ending the episode. A \emph{step} is one model decision and may emit an ordered batch of GUI and CLI actions; step limits and history windows count steps. The \texttt{terminate} action ends the episode with a final answer and a claimed success or failure.

\subsection{Online Agentic RL}
\label{sec:prelim:online}
\label{sec:prelim:opt}

A \emph{rollout} is one attempt at a task: an instance is reset to $s_0$, the policy samples $y_t\sim\pi_\theta(\cdot\mid c_t)$ at each step, and the episode ends on \texttt{terminate} or at the step limit, yielding a trajectory $\tau=(c_1,y_1,\ldots,c_T,y_T)$ and a single terminal reward $r$. Online RL maximizes $J(\theta)=\mathbb{E}_{x\sim\mathcal{X}}\,\mathbb{E}_{\tau\sim\pi_\theta(\cdot\mid x)}[r(x,\tau)]$ over a task pool $\mathcal{X}$ with rollouts from the current policy. Unlike training on fixed demonstrations, every update needs fresh rollouts, so instances must be reset, isolated, and served at scale throughout training.

Our main recipe uses GRPO~\citep{grpo}, which replaces a learned value function with a group baseline: each task is attempted $G$ times, and trajectory $i$ with reward $r_i$ receives the advantage
\begin{equation}
\label{eq:grpo-adv}
    A_i=\frac{r_i-\operatorname{mean}(r_{1:G})}{\operatorname{std}(r_{1:G})+\delta},
\end{equation}
with a small $\delta>0$, shared by all of its policy-generated tokens. A group whose attempts all receive the same reward has zero advantage and contributes no gradient, so rollouts are best spent on tasks that the policy sometimes solves (Section~\ref{sec:training}).

\section{\sys{}}
\label{sec:maigym}

\begingroup
\clubpenalty=10000
\widowpenalty=10000
\brokenpenalty=10000

Online RL for GUI agents usually requires building a separate system for each platform: environments that reset reliably at scale, an agent loop for every benchmark, a pool of tasks with trustworthy rewards, and a trainer for long multimodal trajectories. \sys{} replaces these with one open codebase, organized around a single rollout path that evaluation, data collection, and training all share.

We describe \sys{} in two parts. The first covers its foundation: the system architecture (Section~\ref{sec:arch}) and the mechanisms that make rollouts scalable and efficient (Section~\ref{sec:elastic}). The second follows the improvement loop of Figure~\ref{fig:teaser}, in which \sys{} evaluates the model (Section~\ref{sec:observability}), synthesizes environments and tasks that target its weaknesses (Section~\ref{sec:taskenv}), and retrains it on them (Section~\ref{sec:training}).

\subsection{System Architecture}
\label{sec:arch}

Figure~\ref{fig:arch} shows \sys{}'s architecture and its two design choices. First, all workloads share one rollout path and one agent harness, so evaluation, data collection, and training see the same agent (left). Second, agents reach every environment only through the environment manager, which exposes a common interface and hides environment and deployment details (center). Deployment backends host mobile, computer, and web instances (right), and a dashboard reports system status (bottom).

\begin{figure}[!t]
\centering
\includegraphics[width=\textwidth]{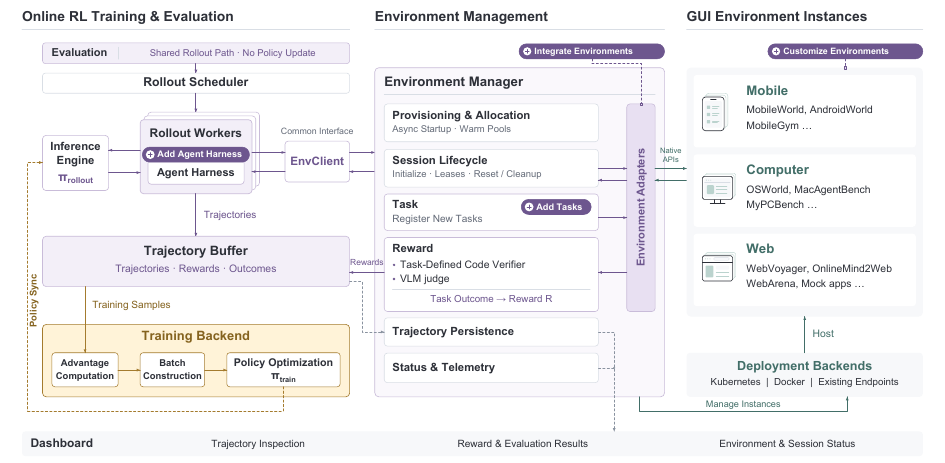}
\caption{\textbf{Architecture of \sys{}.} Left: evaluation and RL training share one harness and rollout path; data collection uses the same path without the policy update. Center: the environment manager provides a common interface. Right: mobile, computer, and web instances hosted by deployment backends. Bottom: the dashboard. Purple tags mark extension points.}
\label{fig:arch}
\end{figure}

\paragraph{One Harness, One Rollout Path.}
The same model can score differently under different harnesses. When evaluation, data collection, and training each keep their own copy of the agent, these copies drift apart in screenshot preprocessing, history truncation, prompt templates, or action parsing. If the training copy keeps the last five screenshots and the evaluation copy keeps ten, we evaluate a different agent from the one we trained. We repeatedly saw such drift as unexplained gaps between training reward and evaluation score. \sys{} therefore keeps one harness for each environment and agent configuration and runs it on a single rollout path. The rollout scheduler queues task attempts. For each attempt, a rollout worker obtains an instance through the EnvClient and runs the harness, which queries the inference engine at every step; the worker then writes the trajectory and its verifier reward to the trajectory buffer. The three workloads differ only in how they use these trajectories: \emph{evaluation} scores and aggregates them, \emph{data collection} stores them, for example as supervised fine-tuning data, and \emph{RL training} uses them to update the policy and then synchronizes the new weights to the inference engine. A fix to the harness therefore reaches all three workloads at once.

\paragraph{Environment Manager.}
The environment manager exposes five operations: request instances, observe, reset to a task, step, and evaluate (Appendix~\ref{app:interface}). Behind these operations, it provisions and allocates instances, manages each session from initialization through reset and cleanup, registers tasks, computes rewards with code verifiers or VLM judges, persists trajectories, and reports status to the dashboard. Environment adapters translate the operations into each environment's native APIs and keep the benchmark's own task initialization and verification; nothing in an adapter refers to RL. Deployment backends host instances on Kubernetes or Docker, or connect to existing phones, computers, and browser sessions.

\paragraph{Extension Points.}
The purple tags in Figure~\ref{fig:arch} mark four places where users extend \sys{}:
\begin{itemize}[leftmargin=*,topsep=0.4em,itemsep=0.35em,parsep=0pt]
    \item \textbf{Add Agent Harness.} A new harness defines context construction, policy invocation, output parsing, and action dispatch, and all three workloads use it unchanged.
    \item \textbf{Add Tasks.} A new task supplies an instruction, required assets, a setup procedure, and a verifier for an existing environment; the synthesis pipeline of Section~\ref{sec:taskenv} creates such tasks at scale.
    \item \textbf{Integrate Environments.} An adapter maps an external benchmark's initialization, observation, action execution, verification, and cleanup to the environment manager's operations. The benchmark keeps its native behavior and gains \sys{}'s instance management and rollout path.
    \item \textbf{Customize Environments.} An environment specification lists an environment's applications, supporting services, and launch and restore configuration. Adding an app to \mobileworld{}, for example, creates a new variant that reuses the existing adapter and deployment backend.
\end{itemize}

\subsection{Scalable and Efficient Rollouts}
\label{sec:elastic}

GUI rollouts cost far more than text rollouts in three ways. Each attempt needs its own phone or desktop in an exact initial state; a training run keeps thousands of such instances busy at once, and some of them fail; and each attempt makes tens of multimodal inference calls before the verifier returns a single reward. \sys{} provides reliable resets, scalable provisioning, and efficient multi-turn inference.

\paragraph{Reliable Task Resets.}
Because every attempt must start from exactly the task's initial state $s_0$ and an online RL run resets environments many thousands of times, we adapt each platform's execution stack for this workload. On mobile, the standard Android emulator needs nested virtualization, which most container clusters lack, so we use ReDroid~\citep{redroid} to run each phone as an ordinary container directly on the host kernel; each reset restores a golden \texttt{/data} snapshot with the matching service state. On desktop, most failures come from virtual machines that report ready too early and from concurrent attempts that interfere with each other; we add stricter readiness checks, recover containers that fail mid-attempt, give each instance its own service ports and artifact cache, and wait for applications to finish starting and saving so the verifier does not read a half-written state. On the web, headless Chromium hosts both real websites and locally served apps, including our mocked apps (Section~\ref{sec:taskenv:env}). With these adaptations, a reset, including queueing and task preparation, takes a median of $165$, $37$, and $6$ seconds on Android, desktop virtual machines, and the browser, respectively (Section~\ref{sec:ana:infra}).

\paragraph{Fast Provisioning at Scale.}
We have tested \sys{} with $20{,}000$ instances running in parallel, and $1{,}000$ fresh instances come up in $328.4$ seconds for mobile and $88.9$ seconds for computer (Section~\ref{sec:ana:infra}). Three design choices make this possible. \emph{First}, provisioning is asynchronous: the environment manager starts instances independently of the requests that asked for them and does not hold a thread while it waits on each backend operation. \emph{Second}, warm pools keep initialized instances ready, and an instance accepts work only after its initialization and registration complete. \emph{Third}, screenshots travel through object storage rather than inside control messages, which keeps the manager's traffic small as the fleet grows. At this scale, some instances, workers, and network calls fail in every run, so \sys{} also confines each failure to the session where it occurs: session leases reject actions from replaced workers, action identifiers make a retried action take effect only once, every operation has a deadline, and verifiers run in separate process groups with resource limits. \sys{} records environment failures separately from zero rewards, so infrastructure faults do not enter the training signal.

\paragraph{Efficient Multi-Turn Inference.}
Figure~\ref{fig:packing}a shows how \sys{} builds each prompt's history so that the inference engine can reuse its KV cache. %
\begin{figure}[!t]
\centering
\includegraphics[width=\linewidth]{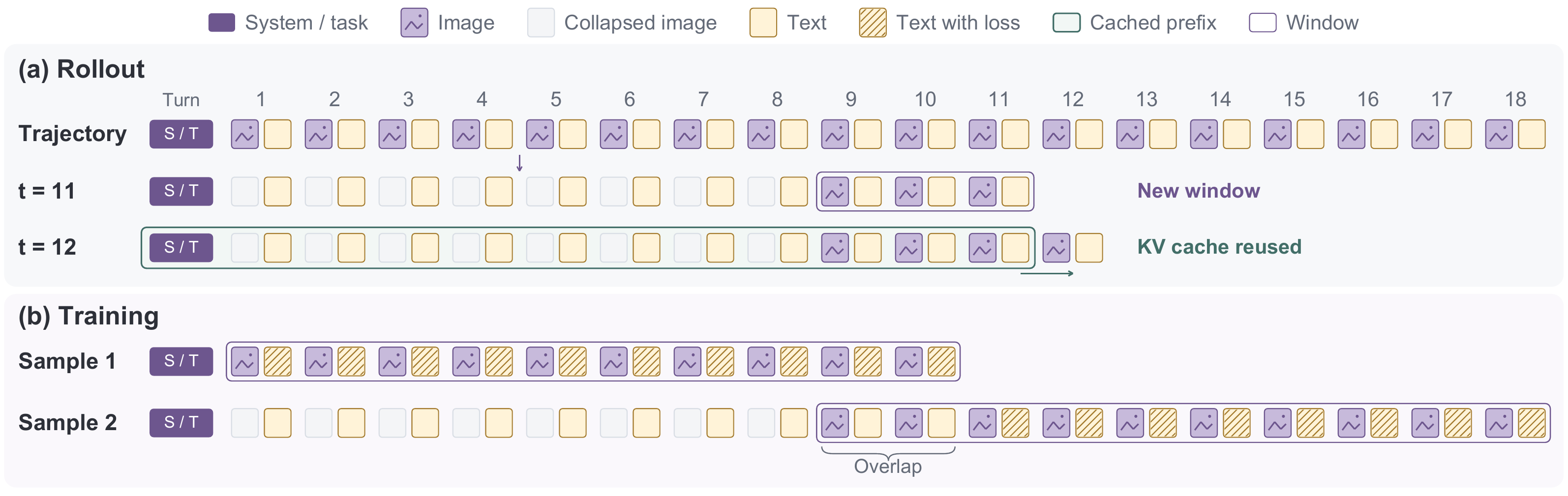}
\caption{\textbf{Window-wise history for rollout and training.} Each turn is one step. With $K=10$ steps per window and $O=2$ overlapping steps, (a)~appending history within a window enables KV-cache reuse; older screenshots become placeholders at window boundaries, while text is retained. (b)~Training reuses these windows, applying loss only to new outputs (hatched). S/T: system prompt and task instruction.}
\label{fig:packing}
\end{figure}%
Consecutive prompts of a trajectory share most of their history, but the engine can reuse cached computation only when a new prompt begins with exactly the tokens of a cached one. The common last-$k$ policy, which keeps the $k$ most recent screenshots, breaks this at every step: dropping the oldest screenshot changes the start of the prompt, so the engine must prefill almost the entire history again. \sys{} instead groups steps into windows of $K$ steps that overlap by $O$ steps. Within a window, new screenshots and text are only appended, so each prompt extends the previous one. At a window boundary, earlier screenshots collapse into placeholders while their text stays. With $K=10$ and $O=2$, turn 11 starts a new window: screenshots 1--8 become placeholders, turns 9--10 carry over, and turn 12 reuses the cached prefix of turn 11. The prefix thus changes once every $K-O$ steps rather than at every step. Because the policy sees between $O+1$ and $K$ recent screenshots, the window size is part of the training recipe, and training reuses the same windows as samples (Figure~\ref{fig:packing}b; Section~\ref{sec:training}).
\sys{} sends the first request of each trajectory to the least-loaded engine and routes the trajectory's later requests to the same engine, which holds its cache. If that engine fails, \sys{} moves the trajectory to a healthy engine, which rebuilds the cache.

\subsection{Evaluation and Diagnostics}
\label{sec:observability}

Evaluation and diagnosis are the first two steps of the improvement loop. \sys{} supports them with three tools: one-command evaluation across benchmarks, a dashboard that traces each score back to its screenshots, and agent-driven diagnosis that turns failed attempts into targets for task synthesis.

\paragraph{One-Command Evaluation.}
A single command evaluates a local checkpoint or an API-served model on any set of registered benchmarks, including the ten in Table~\ref{tab:benchmarks}:
\definecolor{cmdbg}{RGB}{244,247,245}
\definecolor{cmdaccent}{RGB}{61,122,106}
\definecolor{cmdflag}{RGB}{47,110,94}
\definecolor{cmdmuted}{RGB}{128,136,133}
\lstdefinestyle{gymcmd}{basicstyle=\def\ttdefault{zi4}\ttfamily\footnotesize,
  columns=fixed, basewidth=0.5em, keepspaces=true, showstringspaces=false,
  aboveskip=0pt, belowskip=0pt,
  alsoletter={-}, morekeywords={--model,--benchmarks,--trials,--config},
  keywordstyle=\color{cmdflag},
  morekeywords=[2]{mai-gym,eval,train}, keywordstyle=[2]\bfseries,
  moredelim=[s][\color{cmdmuted}]{<}{>},
  literate={\$}{{\textcolor{cmdmuted}{\$}}}1}
\begin{tcblisting}{listing only, enhanced, breakable=false,
    colback=cmdbg, frame hidden, arc=2pt,
    borderline west={2.2pt}{0pt}{cmdaccent},
    left=10pt, right=6pt, top=4pt, bottom=4pt,
    before skip=7pt, after skip=7pt,
    listing options={style=gymcmd}}
$ mai-gym eval --model <checkpoint or API endpoint> \
      --benchmarks mobileworld androidworld mobilegym \
                   osworld-verified osworld-2 macagentbench mypcbench \
                   webvoyager onlinemind2web webarena \
      --trials 3
\end{tcblisting}
The command starts instances in parallel and runs every task through the same harness used in training, so a reported score measures the agent being improved. It preserves each benchmark's step limit, verifier, and native metric, and also reports pass@$n$ and pass$^n$, the fractions of tasks solved in at least one and in all of $n$ trials. When an attempt ends in an environment failure rather than an agent failure, the command reruns it, so infrastructure faults do not lower a model's score; the dashboard still shows every rerun.

\begin{table}[t]
\centering
\captionsetup{font=small,skip=4pt}
\caption{\textbf{Evaluation benchmarks supported by \sys{}.}}
\label{tab:benchmarks}
\footnotesize
\setlength{\tabcolsep}{4pt}
\begin{tabular*}{\linewidth}{@{\extracolsep{\fill}}llccl@{}}
\toprule
Domain & Benchmark & Tasks & Max steps & Metric \\
\midrule
Mobile   & MobileWorld (GUI-only)~\citep{mobileworld} & $117$ & $100$ & Success rate \\
         & AndroidWorld~\citep{androidworld} & $116$ & Task-dependent & Success rate \\
         & MobileGym~\citep{mobilegym} & $256$ & $100$ & binary / partial \\
\midrule
Computer & OSWorld-Verified~\citep{osworld} & $361$ & $100$ & Success rate \\
         & OSWorld 2.0~\citep{osworld2} & $108$ & $500$ & binary / partial \\
         & MacAgentBench~\citep{macagentbench} & $676$ & $50$ & Success rate \\
         & MyPCBench~\citep{mypcbench} & $184$ & $100$ & Perfect-task rate \\
\midrule
Web & WebVoyager~\citep{webvoyager} & $643$ & $15$ & Success rate \\
         & OnlineMind2Web~\citep{onlinemind2web} & $123$ & $50$ & Success rate \\
         & WebArena~\citep{webarena} & $812$ & $30$ & Success rate \\
\bottomrule
\end{tabular*}
\end{table}

\paragraph{Dashboard.}
A score or a reward curve says that an agent fails, not why: a low reward can come from a wrong decision, an unexpected screen, or an emulator that never finished booting, and each cause calls for a different fix. The \sys{} dashboard (Appendix~\ref{app:dashboard}) answers three questions. \emph{Is the fleet healthy?} It shows instance availability, active sessions, and throughput over time. \emph{Which tasks explain a score?} It links each run and benchmark summary to per-task results and to every attempt, including reruns. \emph{Where did an attempt go wrong?} It replays the attempt step by step, aligning each screenshot with the model's output, the executed actions, and the verifier's result, so that agent failures stand apart from environment failures and timeouts.

\paragraph{Agent-Driven Diagnosis.}
Inspecting trajectories by hand does not scale to thousands of attempts across ten benchmarks, so \sys{} hands this work to coding agents. It exposes evaluation records through diagnosis endpoints, which a coding agent such as Claude Code or Codex~\citep{claude_code,codex} queries to retrieve a model's trajectories and rewards. The agent compares successful and failed attempts and analyzes the model's behavior at three levels: individual actions, whole trajectories, and the knowledge that tasks require. The task generator takes the resulting report as input and composes tasks that exercise the weak capabilities (Section~\ref{sec:taskenv:task}), so the model's own failures decide what it trains on next. \sys{} supports running this loop repeatedly; this paper does not yet evaluate repeated rounds.

\subsection{Synthesizing Environments and Tasks}
\label{sec:taskenv}

Diagnosis tells \sys{} what a model needs to practice; synthesis supplies the practice. \sys{} synthesizes every component of an executable task $x=(\mathcal{E},s_0,g,V)$: mocked apps provide new environments $\mathcal{E}$ (Section~\ref{sec:taskenv:env}), cross-app world construction supplies coherent initial states $s_0$ (Section~\ref{sec:taskenv:world}), and task and verifier synthesis derives instructions $g$ and tested verifiers $V$ (Section~\ref{sec:taskenv:task}). Each stage is an agent skill, a set of instructions that a coding agent such as Claude Code follows, and we release the skills together with the apps and tasks they produce.

\subsubsection{Environment Synthesis}
\label{sec:taskenv:env}

Environment synthesis builds \emph{mocked apps}: self-contained applications with resettable backends, so that a policy can repeat a workflow thousands of times without side effects. Mocked apps come in two kinds, depending on whether a real app serves as the reference.

\begin{figure}[!t]
\centering
\includegraphics[width=\linewidth]{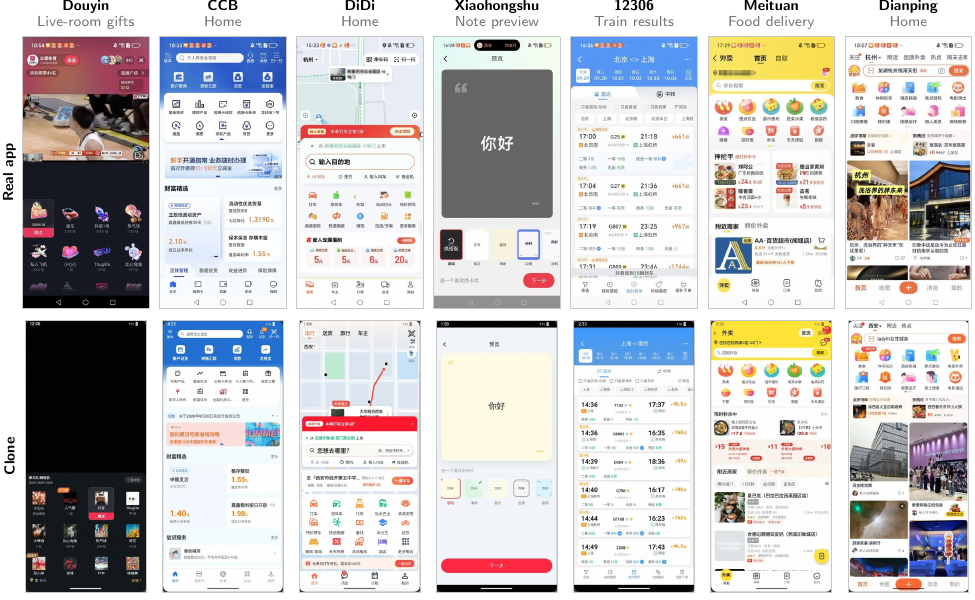}
\caption{\textbf{Real apps and their \sys{} clones.} Top: real-app screens.
Bottom: corresponding clones with independent data. CCB: China Construction
Bank.}
\label{fig:mocked-apps}
\end{figure}

\paragraph{Real-World Clones.}
Many everyday apps act on the real world: they move money, place orders, or publish posts, and these actions cannot be undone, so a policy cannot practice in them thousands of times. We therefore build \emph{real-world clones} of seven such apps, spanning travel (12306), banking (China Construction Bank), ride hailing (DiDi), food delivery (Meituan), local services (Dianping), and social media (Douyin and Xiaohongshu). A coding agent builds each clone from scratch with the real app as a reference. The clone reproduces the app's main user flows, such as booking, ordering, and payment, over a backend that resets between attempts. Testers check each clone against the real app by carrying out the same tasks in both and comparing the screens along the way (Appendix~\ref{app:mockedapps}); Figure~\ref{fig:mocked-apps} shows one screen of each clone next to the real app. The task pipeline below supplies their tasks, and Section~\ref{sec:ana:mocked} shows that training in the 12306 clone improves success in the real app on a physical phone.

\paragraph{Novel Apps.}
Without a real-app reference, we generate \emph{novel apps}: apps invented outright, which no model can have seen during training, for studying whether an agent can learn unfamiliar interfaces through interaction. A coding agent defines each app's initial data, action rules, screens, tasks, and executable checks on a shared HTML5 template. Each novel app is a deterministic state machine that runs in Chromium: the same initial state and action sequence always produce the same result, so resets and rewards are exact. The pipeline tests each app's logic without a browser, runs the same code in Chromium, and validates the app with reference solutions, reset tests, and checks that both runs agree (Appendix~\ref{app:virtualapps}). Section~\ref{sec:ana:novel} reports RL results on these apps.

We release $49$ mocked apps: the seven real-world clones and $42$ novel apps with $840$ tasks. Users can apply the same skills to build apps for their own needs; any such app connects to \sys{} through the web adapter and reuses its evaluation, rollout, and training workflows.

\subsubsection{Cross-App World Construction}
\label{sec:taskenv:world}

Realistic tasks need an initial state that already holds the evidence and resources required to complete them, so the pipeline builds a coherent world before it writes any task (Figure~\ref{fig:taskenv}a). A persona defines the user's identity, role, and goals; a scenario specifies a timeline, events, and constraints. From these, the pipeline generates files and app records that share people, events, and facts: a project brief, a budget spreadsheet, an email thread, and a calendar entry can describe different parts of the same project.

\begin{figure}[!t]
\centering
\includegraphics[width=\linewidth,height=0.45\linewidth]{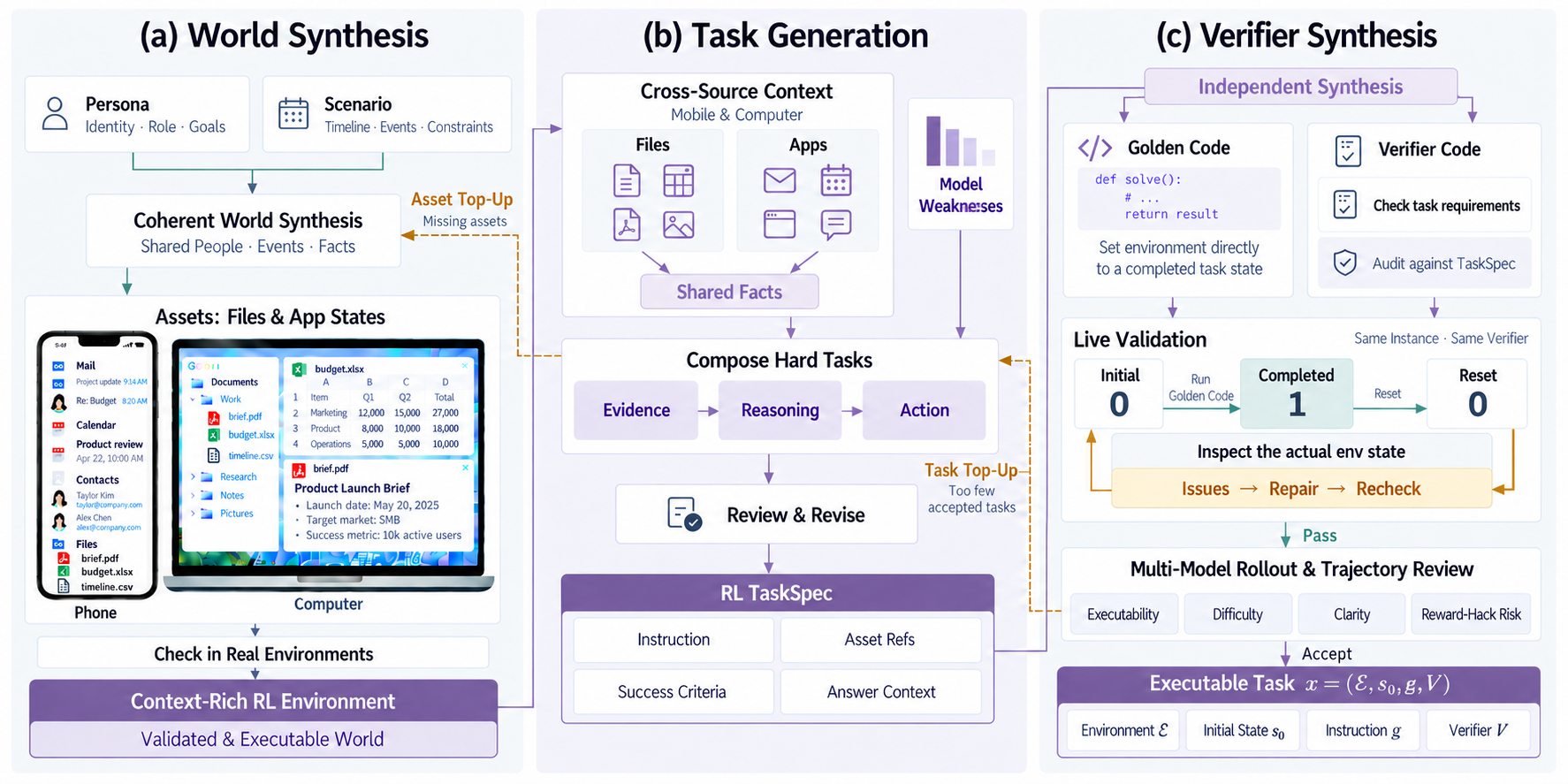}
\caption{\textbf{World, task, and verifier synthesis.} (a)~Construct a coherent world of files and app records. (b)~Generate and review tasks targeting model weaknesses. (c)~Independently synthesize golden and verifier code, validate by execution, and review model rollouts before acceptance.
}
\label{fig:taskenv}
\end{figure}

This shared context lets tasks require finding and combining evidence, as real requests do. Instead of ``create an event named \emph{Budget review}'', a task can ask the agent to move the budget review to the day after the revised spreadsheet is due, which requires reading an email, a spreadsheet, and a calendar. The task is grounded in the world's evidence rather than spelled out as an isolated navigation instruction. The pipeline loads the generated content into a running environment through reusable initialization templates, populating app data on mobile and a desktop workspace on computers, and checks that the expected files and records are present. When a later task needs assets that the world lacks, the pipeline adds them and validates the updated world. The resulting initialization procedure is the task's reproducible $s_0$.

\subsubsection{Task and Verifier Synthesis}
\label{sec:taskenv:task}

Online RL consumes tasks quickly: every task is attempted many times, and a task that the policy always or never solves stops providing gradient. Earlier synthesis either relies on model judges whose false positives a policy can learn to exploit~\citep{zerogui,autoplay} or verifies with code only in simplified applications~\citep{guigenesis,autowebworld}. Our pipeline combines explicit task specifications with execution-based validation inside full \mobileworld{}, OSWorld, and real-world-clone environments (Figure~\ref{fig:taskenv}b--c).

\paragraph{Grounded, Failure-Driven Task Generation.}
The generator extracts facts that the world's files and apps share and composes tasks that chain evidence, reasoning, and action: the agent must find relevant information, work out what to do, and make the requested changes. Diagnosis reports, which summarize the model's failed rollouts, tell the generator which capabilities the new tasks should exercise, so the task pool grows where the model is weak. A review step removes repetitive tasks, clarifies ambiguous instructions, and checks that each requested outcome is meaningful and supported by the world.

\paragraph{Requirements before Verifier.}
Each task specification contains the instruction, references to required assets, success criteria, and an internal answer context that records the supporting facts, calculations, and expected values; the policy never sees the answer context. The specification thus fixes the requirements from the user's goal and the world's evidence before any verifier exists. Two independent synthesis calls then generate \emph{golden code}, which sets the environment directly to a completed state, and \emph{verifier code}, which checks file changes, computed values, and app updates; neither call reads the other's output. A reward audit compares the verifier with the specification to catch missing requirements, unsupported conditions, and wrong expected values, so the verifier checks what the user asked for rather than what one solution happened to do.

\paragraph{Execution before Acceptance.}
The pipeline runs each verifier on a live instance through an initial--completed--reset sequence: the verifier must score $0$ on the initial state, $1$ after the golden code runs, and $0$ again after reset. The pipeline also inspects the actual environment state against the requirements, repairs any issue it finds, and drops candidates that still fail. Because golden code sets the completed state directly, it does not show that an agent can solve the task through the GUI. Several models therefore attempt each remaining task through the agent interface, and a review of their trajectories checks executability, difficulty, instruction clarity, and reward-hacking risk. If too few tasks pass, the pipeline generates more candidates. Models help build and review tasks; online RL rewards come only from verifier execution.

Accepted tasks include the four components of $x$: an environment, reproducible initial state, grounded instruction, and tested verifier. We release $2{,}000$ tasks, $1{,}000$ each for \mobileworld{} and OSWorld; Appendix~\ref{app:taskexamples} shows examples.

\subsection{Online RL Training}
\label{sec:training}

Like evaluation, online RL starts with one command. A YAML file specifies the policy checkpoint, task pools and environments, the RL algorithm and its hyperparameters, rollout budgets and concurrency, the history window, and the training backend and parallelism (Appendix~\ref{app:config}).
\begin{tcblisting}{listing only, enhanced, breakable=false,
    colback=cmdbg, frame hidden, arc=2pt,
    borderline west={2.2pt}{0pt}{cmdaccent},
    left=10pt, right=6pt, top=4pt, bottom=4pt,
    before skip=7pt, after skip=7pt,
    listing options={style=gymcmd}}
$ mai-gym train --config configs/mai-gym-9b.yaml
\end{tcblisting}
Each round, the command collects rollouts on the shared rollout path, computes verifier rewards, updates the policy, and synchronizes the new weights to the inference engines. The trainer builds on slime~\citep{slime}, which trains with Megatron-LM~\citep{megatron} and generates rollouts with SGLang~\citep{sglang}. It supports GRPO~\citep{grpo}, DAPO~\citep{dapo}, GSPO~\citep{gspo}, REINFORCE++~\citep{reinforcepp}, SAO~\citep{sao}, and PPO~\citep{ppo}, which all consume the same records: trajectories, their rewards, and the log-probabilities of policy-generated tokens. Our main recipe uses GRPO, whose objective Appendix~\ref{app:hparams} gives; asynchronous optimizers such as SAO instead update from individually completed trajectories, and Section~\ref{sec:ana:algo} compares the two.

GUI trajectories make three parts of training harder than in text RL. Rollouts are expensive, so they must go to tasks that still teach the policy; trajectories are long and full of screenshots, so training must not re-encode the same history again and again; and the trainer must see exactly the inputs that the inference engine saw. \sys{} addresses these with policy-adaptive task sampling, window-wise packing, and train--rollout consistency checks.

\paragraph{Policy-Adaptive Task Sampling.}
Task synthesis decides \emph{which} tasks exist; task sampling decides \emph{how often} the policy attempts each one. GRPO learns only from groups whose attempts receive different rewards, so rollouts spent on tasks that the policy always or never solves are wasted. Before each rollout round, the scheduler estimates each task's success rate from its past attempts. Unseen tasks and tasks that the policy sometimes solves receive the full group of $G$ attempts. Tasks that the policy has never solved, and tasks that it has mastered with a success rate above $\eta$, receive small but nonzero budgets, so the scheduler rechecks them as the policy changes; we use $\eta=0.8$ and budgets of $4$ and $2$ attempts, respectively. The trainer computes a task's advantages only after all of its attempts are resolved, so completion order and trajectory length do not bias its baseline.

\paragraph{Window-Wise Packing.}
Figure~\ref{fig:packing}b shows how training reuses the rollout windows: each window of a trajectory becomes one causal training sequence, so the trainer encodes a screenshot once per window rather than once per step. The two common alternatives each pay a cost. \emph{Step-wise} training turns every step into a separate sample and re-encodes the shared history each time. \emph{Trajectory-level} training packs a whole trajectory into one sequence, which is exact only if each context extends the previous one and quickly exceeds accelerator memory on long, screenshot-heavy trajectories. Windows sit between the two. Loss applies only to each window's new steps: with $K=10$ and $O=2$, the first sample trains turns 1--10 and the second trains turns 11--18, using turns 9--10 only as context, so no output is trained twice. Each window replays the exact tokens sampled during rollout, and the packer falls back to step-wise samples for any trajectory whose turns do not form a consistent token prefix. To fit long sequences in memory, screenshots reach the trainer as compressed images and are decoded only when their microbatch is scheduled, while selective activation recomputation in the vision tower and configurable context, pipeline, and data parallelism keep memory bounded.

\paragraph{Train--Rollout Consistency.}
Training is only as correct as its agreement with rollout: if the trainer rebuilds a prompt or processes an image differently from the inference engine, the policy is updated on inputs it never saw. Launch-time checks catch known faults in multimodal processing. During the run, \sys{} compares trainer and rollout log-probabilities on every response token and gates the run on the $k_3$ estimator of their KL divergence. Unlike the mean absolute difference, $k_3$ is dominated by the few tokens that disagree sharply, so a sudden rise points to a corrupted generation kernel, a misaligned packed history, or a version skew between trainer and engine.

\endgroup

\Needspace{5\baselineskip}
\section{Experiments}
\label{sec:exp}

\subsection{Experimental Settings}
\label{sec:exp_settings}

\paragraph{Training MAI-Gym-9B.}
We train MAI-Gym-9B from Qwen3.5-9B~\citep{qwen35_9b} in two stages. The warm-up stage applies supervised fine-tuning on data that covers both mobile and desktop environments, so that the model learns to follow mixed GUI and CLI instructions in both domains.

The RL stage then trains this initialization on mobile and computer tasks \emph{jointly}, in a single run. A common recipe instead runs RL separately in each domain and merges the resulting experts by parameter averaging~\citep{uitars2}. Because \sys{} serves phones and desktops through one rollout path, a run draws tasks from both domains, training one policy rather than reconciling separate experts afterward. We train with GRPO~\citep{grpo} on $5{,}000$ tasks from the synthesis pipeline (Section~\ref{sec:taskenv:task}), mixed across \mobileworld{} and OSWorld environments, with rewards from their code verifiers only, and use policy-adaptive task sampling and window-wise history with $K=10$ and $O=2$ (Section~\ref{sec:training}).
The run is launched by \texttt{mai-gym train} with the configuration in Appendix~\ref{app:config}, using $512$ GPUs and a peak of $20{,}000$ concurrent environments, and trains for approximately $96$ hours.
Section~\ref{sec:ana:joint} compares joint training with merging per-domain experts; Appendix~\ref{app:hparams} lists the main hyperparameters.

\paragraph{Benchmarks and Metrics.}
We evaluate MAI-Gym-9B on seven benchmarks, three for mobile use and four for computer use: \mobileworld{}~\citep{mobileworld}, AndroidWorld~\citep{androidworld}, MobileGym~\citep{mobilegym}, OSWorld-Verified~\citep{osworld}, OSWorld 2.0~\citep{osworld2}, MacAgentBench~\citep{macagentbench}, and MyPCBench~\citep{mypcbench}.
\mobileworld{} and OSWorld-Verified are \emph{Task OOD}: they share their environments, but not their tasks, with training. The other five are \emph{Env OOD}: MAI-Gym-9B never trains in the environments of AndroidWorld, MobileGym, MacAgentBench, or MyPCBench, and OSWorld 2.0 adds $31$ self-hosted websites and long-horizon workflows that our training environments do not contain.

Our evaluations run through the \texttt{mai-gym eval} command, with the agent harness used in training. We report success rates under each benchmark's own evaluation metric (Table~\ref{tab:benchmarks}). For OSWorld 2.0, we report binary accuracy / partial score, where partial score is the mean task reward in percent; for MobileGym, we report the partial score. To summarize results, we use unweighted averages: Mobile Avg and Computer Avg over the three mobile and four computer benchmarks, Cross Avg as the mean of the two, and Task-OOD Avg and Env-OOD Avg over the two Task OOD and five Env OOD benchmarks; all averages use the binary score of OSWorld 2.0 and the partial score of MobileGym.
In Table~\ref{tab:main}, parenthesized values give gains over Qwen3.5-9B in percentage points; dashes denote unreported results.

\paragraph{Baselines.}
We compare MAI-Gym-9B with three groups of models (Table~\ref{tab:main}). \emph{Frontier models}: Claude Opus 4.8~\citep{opus48}, GPT-5.5~\citep{gpt55}, and Qwen3.8-Max~\citep{qwen38_release}, all accessed through APIs, together with the open-weight Qwen3.8-27B~\citep{qwen38_27b_modelcard}. \emph{GUI agent models}: GUI-Owl-1.5-8B and GUI-Owl-1.5-32B~\citep{mobileagent35}, UI-Venus-2-9B~\citep{uivenus2}, SCALECUA-Qwen3.5-9B~\citep{scalecua_rl}, EvoCUA-1.5-32B~\citep{evocua15}, CUA-GYM-35B-A3B and CUA-GYM-397B-A17B~\citep{cuagym}, and Qwen-CUA~\citep{qwencua}; UI-Venus-2-9B and SCALECUA-Qwen3.5-9B match MAI-Gym-9B in size. \emph{Base model}: the Qwen3.5-9B backbone, which measures the gain of the whole recipe. Baseline entries come from published reports and leaderboards, except eight entries from our runs: Qwen3.8-Max on MacAgentBench and MyPCBench, Qwen3.8-27B and Qwen3.5-9B on MacAgentBench, and both GUI-Owl sizes on MyPCBench and OSWorld 2.0.

\subsection{Main Results}
\begin{table*}[t]
\centering
\caption{\textbf{Mobile and computer benchmark results (\%).} \emph{Task OOD}: unseen tasks in training environments; \emph{Env OOD}: unseen environments.}
\label{tab:main}
\scriptsize
\setlength{\tabcolsep}{1.2pt}
\begin{tabular*}{\textwidth}{@{\extracolsep{\fill}}lccccccccc@{}}
\toprule[1.2pt]
\multirow{4}{*}{\textbf{\textsc{Model}}} &
\multirow{4}{*}{\textbf{\textsc{Size}}} &
\multirow{4}{*}{\textbf{\textsc{Access}}} &
\multicolumn{3}{c}{\textbf{\textsc{Mobile}}} & \multicolumn{4}{c}{\textbf{\textsc{Computer}}} \\
\cmidrule(lr){4-6}\cmidrule(lr){7-10}
& & & \textit{Task OOD} & \multicolumn{2}{c}{\textit{Env OOD}} & \textit{Task OOD} & \multicolumn{3}{c}{\textit{Env OOD}} \\
\cmidrule(lr){4-4}\cmidrule(lr){5-6}\cmidrule(lr){7-7}\cmidrule(lr){8-10}
& & & \makecell{\mobileworld{}\\GUI-only} & \makecell{Android\\World} & \makecell{Mobile\\Gym} & \makecell{OSWorld\\Verified} & \makecell{OSWorld\\2.0} & \makecell{MacAgent\\Bench} & \makecell{MyPC\\Bench} \\
\midrule
\rowcolor{gray!15}
\multicolumn{10}{l}{\textit{Frontier models}}\\
Claude Opus 4.8
    & -- & API
    & 69.2 & 75.0 & 50.7
    & 83.4 & 20.3/54.8 & 58.4 & 62.0 \\
GPT-5.5
    & -- & API
    & 62.4 & 74.1 & 48.1
    & 78.7 & 13.9/47.5 & 66.7 & 47.3 \\
Qwen3.8-Max
    & 2.4T-A95B & API
    & 77.8 & 85.3 & 61.3
    & 86.1 & 19.4/46.7 & 74.3 & 48.6 \\
Qwen3.8-27B
    & 27B & Open
    & 70.1 & 81.9 & 57.5
    & 84.3 & 19.4/48.0 & 65.4 & 28.5 \\
\rowcolor{gray!15}
\multicolumn{10}{l}{\textit{GUI agent models}}\\
GUI-Owl-1.5-8B
    & 8B & Open
    & 38.2 & 71.6 & 15.1
    & 52.9 & 0.0/1.6 & 10.2 & 1.8 \\
GUI-Owl-1.5-32B
    & 32B & Open
    & 43.9 & 69.8 & 20.3
    & 56.5 & 0.9/2.5 & 17.2 & 6.3 \\
UI-Venus-2-9B
    & 9B & Open
    & 75.2 & 80.2 & 52.7
    & 70.8 & 0.0/7.5 & 36.2 & 6.3 \\
SCALECUA-Qwen3.5-9B
    & 9B & Open
    & 12.8 & 47.4 & 13.5
    & 68.7 & 0.0/2.2 & 3.9 & 1.1 \\
Qwen-CUA
    & 397B-A17B & Open
    & -- & -- & --
    & 86.2 & 18.5/48.4 & 69.2 & 58.7 \\
EvoCUA-1.5-32B
    & 32B & Unreleased
    & -- & -- & --
    & 63.2 & --/-- & -- & -- \\
CUA-GYM-35B-A3B
    & 35B-A3B & Unreleased
    & -- & -- & --
    & 62.1 & --/-- & -- & -- \\
CUA-GYM-397B-A17B
    & 397B-A17B & Unreleased
    & -- & -- & --
    & 72.6 & --/-- & -- & -- \\
\midrule
\rowcolor{gray!15}
\multicolumn{10}{l}{\textit{Ours}}\\
Qwen3.5-9B (base model)
    & 9B & Open
    & 18.0 & 57.8 & 9.0
    & 41.8 & 0.0/2.5 & 20.7 & 6.5 \\
\fullrowcolor{mairow}
    & &
    & \textbf{78.4} & \textbf{75.9} & \textbf{46.5}
    & \textbf{76.2} & \textbf{3.7/11.3} & \textbf{41.3} & \textbf{15.2} \\
\fullrowcolor{mairow}
\multirow{-2}{*}{\textbf{MAI-Gym-9B}}
    & \multirow{-2}{*}{\textbf{9B}} & \multirow{-2}{*}{\textbf{Open}}
    & {\tiny\textbf{(+60.4)}} & {\tiny\textbf{(+18.1)}} & {\tiny\textbf{(+37.5)}}
    & {\tiny\textbf{(+34.4)}} & {\tiny\textbf{(+3.7/+8.8)}} & {\tiny\textbf{(+20.6)}} & {\tiny\textbf{(+8.7)}} \\
\bottomrule[1.2pt]
\end{tabular*}
\end{table*}
\paragraph{Task OOD Benchmarks.}
Table~\ref{tab:main} evaluates one MAI-Gym-9B checkpoint on all seven benchmarks, whereas four of the GUI agent baselines report computer use only. On the two benchmarks whose environments appear in training, MAI-Gym-9B reaches $78.4\%$ on \mobileworld{} and $76.2\%$ on OSWorld-Verified, the highest scores among GUI agent models of comparable size. It leads UI-Venus-2-9B by $3.2$ and $5.4$ points and GUI-Owl-1.5-8B by $40.2$ and $23.3$ points, respectively. It also outperforms the larger GUI agents that report these benchmarks: GUI-Owl-1.5-32B on both, and EvoCUA-1.5-32B, CUA-GYM-35B-A3B, and CUA-GYM-397B-A17B on OSWorld-Verified; only Qwen-CUA, a 397B-A17B model, scores higher on OSWorld-Verified. Against frontier models, MAI-Gym-9B surpasses Claude Opus 4.8 ($69.2\%$), GPT-5.5 ($62.4\%$), and Qwen3.8-Max ($77.8\%$) on \mobileworld{}, and comes within $2.5$ points of GPT-5.5 on OSWorld-Verified. Because no benchmark task is used in training, these scores measure how well the policy solves new tasks in environments it has trained in.

\paragraph{Env OOD Benchmarks.}
The other five benchmarks test whether these abilities carry over to environments where MAI-Gym-9B lacks environment-specific knowledge. MAI-Gym-9B remains competitive on the two mobile Env-OOD benchmarks, achieving $75.9$ on AndroidWorld and $46.5$ on MobileGym, within $4.3$ and $6.2$ points of UI-Venus-2-9B and above GUI-Owl-1.5-8B on both. On the more challenging computer Env-OOD benchmarks, it reaches $3.7/11.3$ on OSWorld 2.0, $41.3$ on MacAgentBench, and $15.2$ on MyPCBench, leading all similarly sized open GUI agents with reported results.

\begin{wrapfigure}{r}{0.5\textwidth}
\vspace{-1.2em}
\centering
\includegraphics[width=0.93\linewidth]{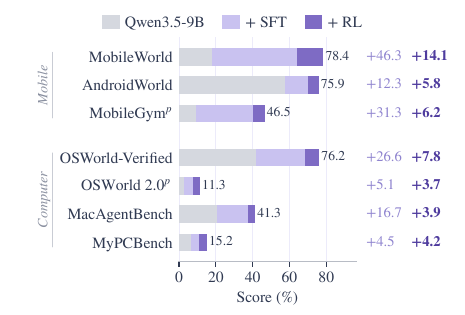}
\caption{\textbf{Gains from SFT warm-up and joint RL.}}
\label{fig:stage_gains}
\vspace{-1.6em}
\end{wrapfigure}

\paragraph{Gains from SFT and Joint RL.}
Figure~\ref{fig:stage_gains} splits each score of MAI-Gym-9B into the reported score of its Qwen3.5-9B base, the gain from the SFT warm-up, and the gain from joint RL, with the SFT warm-up checkpoint evaluated under our protocol. The superscript $p$ marks partial scores for MobileGym and OSWorld 2.0. The SFT warm-up provides most of the improvement: it raises Task-OOD Avg from $29.9$ to $66.4$ and Env-OOD Avg from $18.8$ to $32.3$. Joint RL then raises Cross Avg from $44.1$ to $50.5$, with gains of $8.7$ points on Mobile Avg and $4.2$ on Computer Avg, so training the two domains in one run improves both rather than trading one for the other. RL adds $14.1$ and $7.8$ points on the two Task OOD benchmarks and $4.2$ on Env-OOD Avg, and it improves every Env OOD benchmark, so what the policy learns from synthesized tasks in two environments transfers beyond them.

\begin{figure}[!t]
\centering
\includegraphics[width=\textwidth]{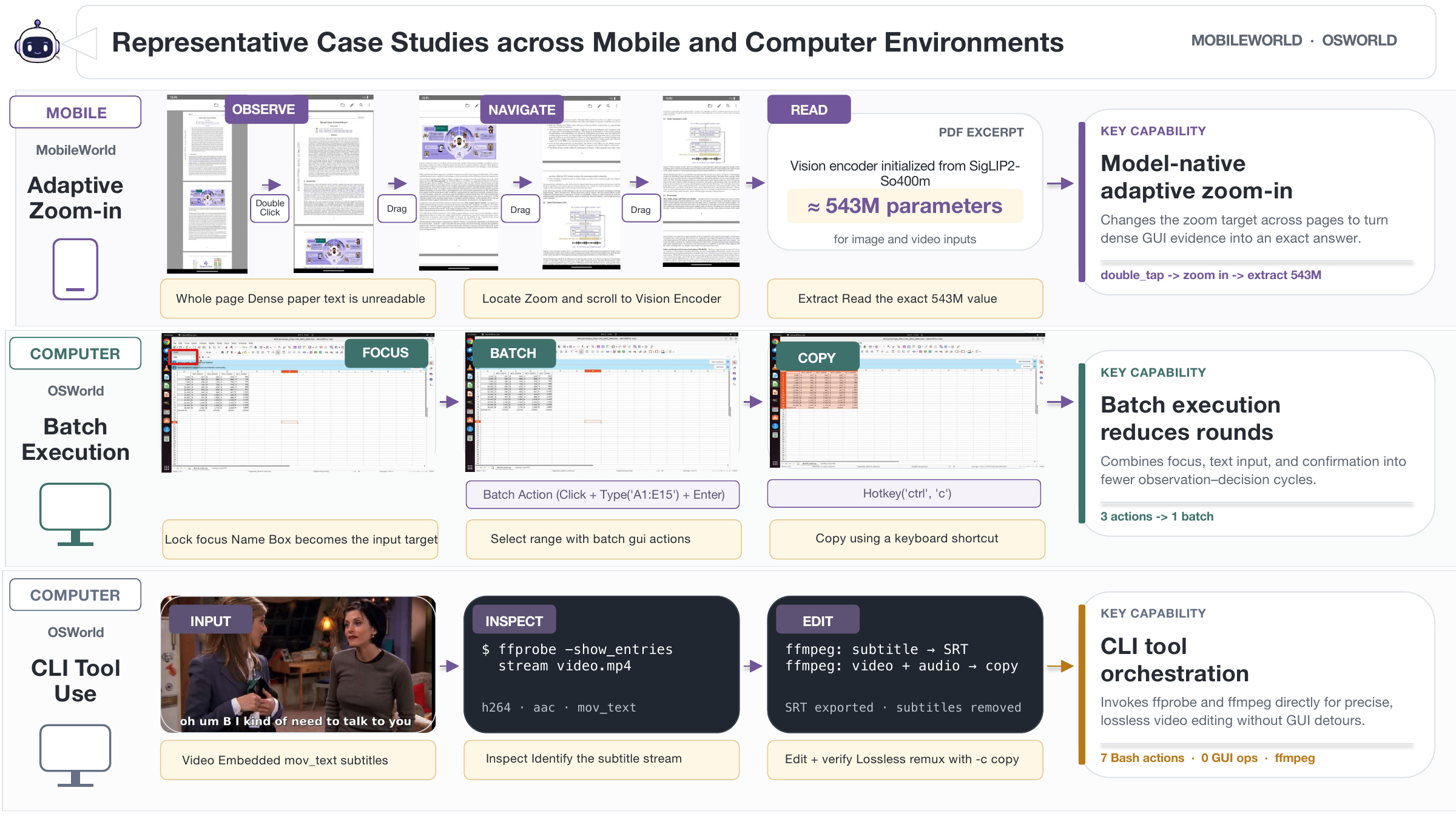}
\caption{\textbf{Efficient behaviors learned by MAI-Gym-9B.} Top: adaptive zoom in \mobileworld{}. Middle: batched GUI actions in OSWorld. Bottom: CLI tool use in OSWorld.}
\label{fig:case_studies}
\end{figure}

\paragraph{Efficient Behaviors Learned through Training.}
Figure~\ref{fig:case_studies} illustrates efficient behaviors that MAI-Gym-9B learns through training. On computer tasks, the model can invoke CLI tools when they are the most efficient interface. In OSWorld, it uses \texttt{ffprobe} to inspect video streams and \texttt{ffmpeg} to export subtitles and remux the file losslessly, instead of taking a longer GUI route for media editing. The model also supports batch execution for GUI actions. In spreadsheet-style interactions, steps such as locking focus, entering a range, confirming the input, and copying with a keyboard shortcut can be combined into fewer higher-level actions, which reduces execution steps, observation--decision cycles, and user token cost. On mobile tasks, MAI-Gym-9B further shows model-native adaptive zoom-in. By using app-native double-tap zoom and shifting the zoom target across pages, it can inspect small details that are unreadable in the full-page view, which helps with information retrieval and exact numerical extraction. These examples suggest that MAI-Gym-9B improves not only by selecting better actions, but also by choosing the right interaction mode for each task, including direct GUI manipulation, batched GUI execution, and external tool use.

\subsection{Analysis}
\label{sec:analysis}

We next analyze \sys{} and its training recipe. Each analysis answers one question: whether the infrastructure runs efficiently at the scale that online GUI RL needs (Section~\ref{sec:ana:infra}), whether joint RL beats merging per-domain experts (Section~\ref{sec:ana:joint}), how RL gains grow with the number of synthesized tasks (Section~\ref{sec:ana:tasks}), whether training in a real-world clone transfers to the real app (Section~\ref{sec:ana:mocked}), whether an agent can learn novel apps from reward alone (Section~\ref{sec:ana:novel}), and how GRPO compares with SAO on \sys{} (Section~\ref{sec:ana:algo}).

\subsubsection{Infrastructure Runs Efficiently at Scale}
\label{sec:ana:infra}
\begin{table}[!t]
\centering
\caption{\textbf{Environment efficiency and reliability.} Bring-up ends when 1,000 environments are GUI-ready. Unscored aborts are attempts rerun after environment failures, without scoring.}
\label{tab:infra}
{\footnotesize
\setlength{\tabcolsep}{4pt}
\begin{tabular*}{\linewidth}{@{\extracolsep{\fill}}lccc@{}}
\toprule
Environment & \makecell{Bring-up time for 1k envs (s)} & Reset p50 / p95 (s) & Unscored aborts (\%) \\
\midrule
Mobile  & 328.4 {\scriptsize ($\geq 6\times$ faster)}\textsuperscript{$\dagger$} & 165.0 / 184.0 & 1.2 \\
Computer     & 88.9 & 37.2 / 100.1 & 1.9 \\
Browser          & 94.2 & 6.0 / 27.0 & 0.0 \\
\bottomrule
\end{tabular*}\par}
\vspace{3pt}
\begin{minipage}{\linewidth}
\scriptsize $\dagger$ Mobile speedup versus estimated KVM/bare-metal: 100 serial batches of 10 with 20-second waits require at least 1,980 seconds.
\end{minipage}

\vspace{10pt}
\caption{\textbf{Efficiency of step-wise and window-wise history.}}
\label{tab:infra_train}
{\footnotesize
\setlength{\tabcolsep}{3pt}
\begin{tabular*}{\linewidth}{@{\extracolsep{\fill}}lcccc@{}}
\toprule
History & \makecell{Prefix-cache hit (\%)} & \makecell{Rollout steps/s} & \makecell{Train input tokens/traj.} & \makecell{Train update (s)} \\
\midrule
Step-wise & 42.5 & 2.3 & 566,375 & 2166.4 \\
\makecell[l]{Window-wise ($K{=}10$, $O{=}2$)} & 85.9\,{\scriptsize(+102.1\%)} & 1.9\,{\scriptsize(-18.9\%)} & 86,855\,{\scriptsize(-84.7\%)} & 810.9\,{\scriptsize(-62.6\%)} \\
\bottomrule
\end{tabular*}\par}
\end{table}

Table~\ref{tab:infra} summarizes the environment efficiency of \sys{}. Across mobile, computer, and browser settings, \sys{} supports low-abort operation while bringing up 1,000 GUI-ready environments within a few minutes. In particular, mobile bring-up takes 328.4 seconds, which is at least $6\times$ faster than our estimate for a serial KVM-style setup, suggesting that the environment manager is effective for scaling online RL in GUI domains. Evaluating a MAI-Gym-9B checkpoint on all seven benchmarks with three trials per task takes approximately three hours.

Table~\ref{tab:infra_train} compares step-wise and window-wise history. Window-wise history yields much higher prefix-cache reuse and substantially lower training token cost, which in turn reduces training update time from 2166.4 to 810.9 seconds, although rollout throughput decreases moderately. Overall, these measurements suggest that window-wise history offers a better efficiency trade-off in our training setup.

\subsubsection{Joint RL Beats Expert Merging}
\label{sec:ana:joint}

We compare joint RL on mobile and computer tasks with a two-stage alternative: training separate per-domain experts and then merging them by weighted parameter averaging~\citep{goddard-etal-2024-arcees}. All runs start from the same warm-up checkpoint, and the joint RL run uses the combined training budget of the two experts.

\begin{figure}[!t]
\centering
\captionsetup{position=top,skip=3pt,font=small,justification=raggedright,singlelinecheck=false}
\begin{minipage}[t]{0.48\linewidth}
\vspace{0pt}
\centering
\captionof{table}{\centering\textbf{Joint RL vs.\ expert merging.}}
\label{tab:crossstage}
\footnotesize
\setlength{\tabcolsep}{3pt}
\begin{tabular*}{\linewidth}{@{\extracolsep{\fill}}lcc@{}}
\toprule
Setting & MobileWorld & OSWorld-Verified \\
\midrule
Mobile expert       & 76.4 & 71.4 \\
Computer expert     & 73.5 & 73.8 \\
\midrule
Merge 70/30         & 74.4 & 68.4 \\
Merge 50/50         & 76.4 & 73.7 \\
Merge 30/70         & 73.5 & 74.8 \\
\midrule
\fullrowcolor{mairow}
\textbf{Joint RL}   & \textbf{78.4} & \textbf{76.2} \\
\bottomrule
\end{tabular*}
\end{minipage}\hfill
\begin{minipage}[t]{0.48\linewidth}
\vspace{0pt}
\centering
\captionof{figure}{\centering\textbf{Synthesized task scaling.}}
\label{fig:taskscale}
\definecolor{figlight}{HTML}{C9C2F0}
\definecolor{figdark}{HTML}{7E6BC4}
\begin{tikzpicture}
\begin{axis}[
    width=\linewidth, height=3.3cm,
    ybar, bar width=12pt,
    ymin=60, ymax=82, ytick={60,65,70,75,80},
    xtick={1,2,3}, xticklabels={0.5k, 1k, 5k},
    enlarge x limits=0.25,
    axis x line*=bottom, axis y line*=left,
    ymajorgrids, grid style={gray!25},
    xlabel={Synthesized training tasks},
    ylabel={Success rate (\%)},
    label style={font=\scriptsize},
    tick label style={font=\scriptsize},
    nodes near coords,
    nodes near coords style={font=\tiny},
    point meta=explicit symbolic,
    legend style={font=\scriptsize, draw=none, fill=none, at={(0.5,1.02)}, anchor=south, legend columns=2, /tikz/every even column/.append style={column sep=6pt}},
    legend image code/.code={\draw[#1] (0cm,-0.08cm) rectangle (0.24cm,0.1cm);},
]
\addplot[fill=figlight, draw=none] coordinates {
    (1,69.2) [69.2]
    (2,73.5) [73.5]
    (3,78.4) [78.4]
};
\addplot[fill=figdark, draw=none] coordinates {
    (1,69.7) [69.7]
    (2,70.9) [70.9]
    (3,76.2) [76.2]
};
\legend{\mobileworld{}, OSWorld-Verified}
\end{axis}
\end{tikzpicture}
\end{minipage}
\par\vspace{6pt}
\begin{minipage}[t]{0.48\linewidth}
\vspace{0pt}
\centering
\captionof{figure}{\centering\textbf{Clone-to-real transfer (12306, 52 tasks).}}
\label{fig:sim2real-sr}
\includegraphics[width=\linewidth]{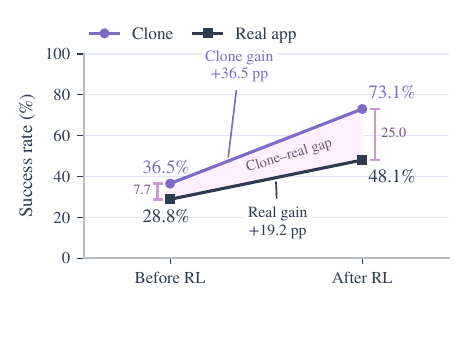}
\end{minipage}\hfill
\begin{minipage}[t]{0.48\linewidth}
\vspace{0pt}
\centering
\captionof{figure}{\centering\textbf{GRPO vs.\ SAO on \mobileworld{}.}}
\label{fig:optimizer}
\includegraphics[width=\linewidth]{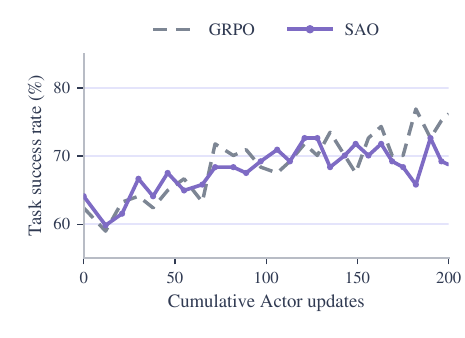}
\end{minipage}
\end{figure}

Table~\ref{tab:crossstage} reports success rates in percent, with merge weights listed in mobile/computer order. The merged models retain a substantial portion of the domain-specific gains from the two experts, but they do not match joint training. No single merge leads in both domains: the best merge for MobileWorld (50/50) reaches 76.4\%, and the best for OSWorld-Verified (30/70) reaches 74.8\%. Joint RL attains 78.4\% and 76.2\%, above the best merge in each domain.
These results suggest that joint RL outperforms specialist merging in both domains.

\begin{figure}[!tp]
\centering
\includegraphics[width=\linewidth]{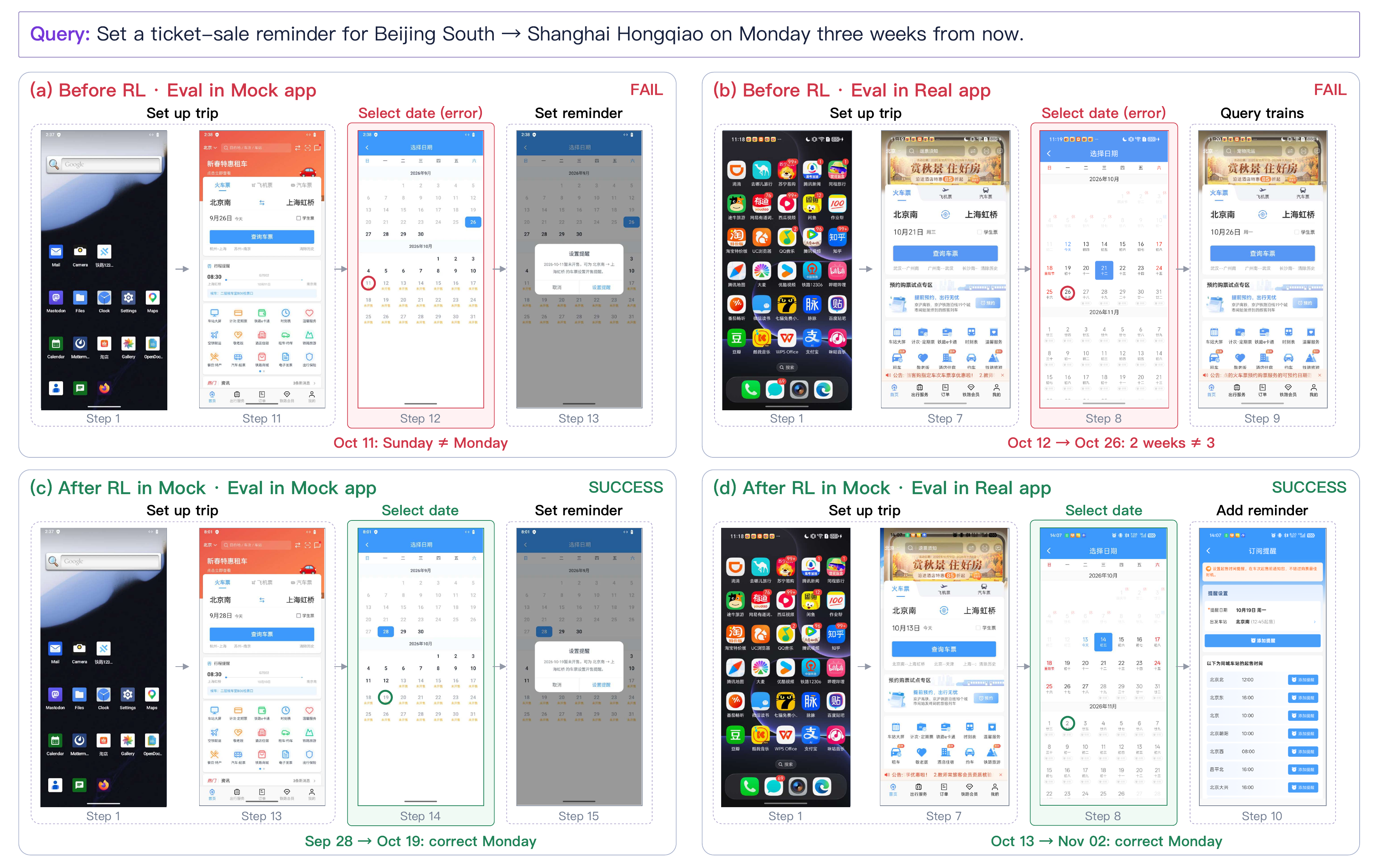}
\caption{\textbf{Relative-date reasoning after clone-only RL.} Top: incorrect dates before RL. Bottom: the correct Monday is selected and the reminder is set in both the clone and real app after RL.}
\label{fig:sim2real-date}
\end{figure}

\subsubsection{Scaling Synthesized Tasks Improves Generalization}
\label{sec:ana:tasks}

We examine how larger synthesized task pools affect performance.
From the synthesized \mobileworld{} and OSWorld tasks, we randomly sample training sets of $500$, $1{,}000$, and $5{,}000$ tasks in total across the two environments. All RL runs use the same algorithm, hyperparameters, and rollout budget.

Figure~\ref{fig:taskscale} shows that scaling synthesized training tasks improves generalization on \mobileworld{} and OSWorld-Verified, whose benchmark tasks are not used in training. Increasing the number of tasks from $500$ to $5{,}000$ improves the success rate from $69.2\%$ to $78.4\%$ on \mobileworld{} and from $69.7\%$ to $76.2\%$ on OSWorld-Verified. With the rollout budget held constant, these results show the benefit of training on more diverse synthesized tasks.

\subsubsection{Clone Training Transfers to the Real App}
\label{sec:ana:mocked}

Using $52$ paired tasks, we compare success rates before and after clone-only RL in the 12306 clone and the real 12306 app on a physical phone. Figure~\ref{fig:sim2real-sr} shows clone-only RL improves success from $36.5\%$ to $73.1\%$ in the clone and from $28.8\%$ to $48.1\%$ in the real app (a gain of $19.2$ percentage points). Figure~\ref{fig:sim2real-date} illustrates shared relative-date reasoning: after RL, the policy selects the correct Monday and sets the reminder in both apps. Transfer remains incomplete: the clone--real gap widens from $7.7$ to $25.0$ percentage points, so clone gains do not fully predict real-app performance.

\subsubsection{Agents Learn Unfamiliar Apps from Reward}
\label{sec:ana:novel}
\label{sec:virtualapps}

\begin{figure}[!tp]
\centering
\includegraphics[width=\linewidth]{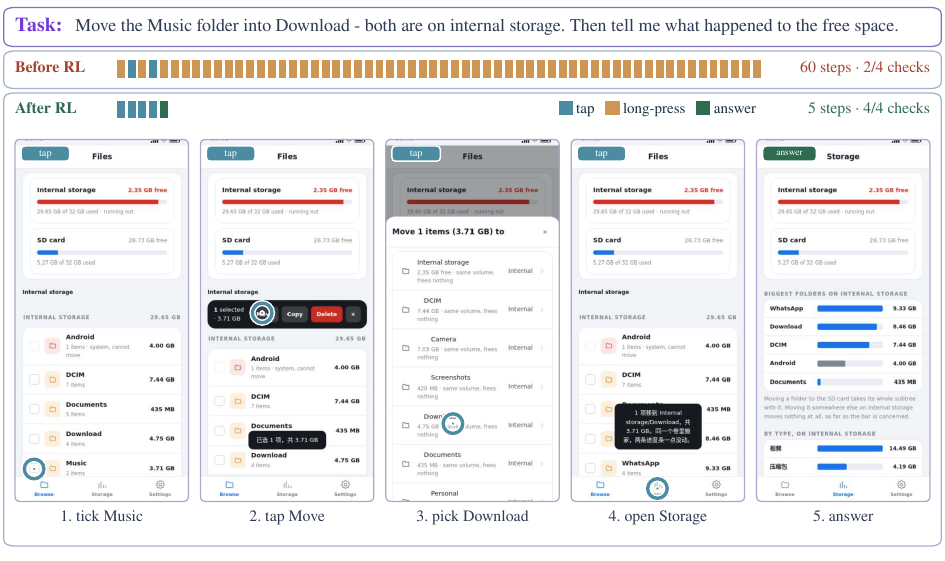}
\caption{\textbf{RL enables efficient task completion in a novel app.} Before RL, repeated ineffective gestures exhaust the $60$-step budget; after RL, the agent moves the folder and reports unchanged free space, passing all four checks in five steps.}
\label{fig:novel-case}
\end{figure}

Novel apps test learning unfamiliar interfaces. We train on $252$ tasks and evaluate on $62$ held-out ones, one attempt each, under a $60$-step budget. RL raises success from $25.8\%$ to $45.2\%$, reduces budget-exhausting attempts from $27$ to $5$, and shortens mean trajectories from $33.3$ to $16.4$ steps. Figure~\ref{fig:novel-case} illustrates one held-out task. RL uses only verifier rewards, with no demonstrations of these apps.

\Needspace{10\baselineskip}
\subsubsection{Comparing GRPO and SAO}
\label{sec:ana:algo}

\sys{} supports RL algorithms through shared environment interfaces, rollout collection, and verifier rewards (Section~\ref{sec:training}). We compare our GRPO recipe with SAO~\citep{sao}, which replaces grouped rollouts with asynchronous single-rollout updates and uses a learned value model for advantage estimation.

Figure~\ref{fig:optimizer} shows similar early task success for both recipes, with SAO ending the displayed interval at a slightly lower success rate than GRPO. In our timing comparison, SAO reduces sampling wait time, but the added critic computation, including two critic updates per actor update, offsets this benefit. Thus, SAO shows no clear overall training-time advantage over GRPO under the tested configurations.

\FloatBarrier

\section{Related Work}
\label{sec:related}

\paragraph{RL for GUI Agents.}
Online RL has become a standard way to improve GUI agents, and most work contributes a training recipe for one domain. On Android, DigiRL, DistRL, MobileRL, and MobileGUI-RL study reward design, asynchronous collection, and adaptive training~\citep{digirl,distrl,mobilerl,mobileguirl}. On the web, WebRL trains with a self-evolving curriculum and WebAgent-R1 with end-to-end multi-turn RL~\citep{webrl,webagentr1}. On computers, ComputerRL scales RL with parallel execution, DART-GUI decouples training from rollouts, and Qwen-CUA learns native computer use from screenshots~\citep{computerrl,dartgui,qwencua}. Later systems refine the training loop: EvoCUA-1.5 adds asynchronous multi-turn RL with policy-aware filtering and an adaptive curriculum, UI-Mate combines environment-grounded training with in-context demonstrations, and SCALECUA couples verifiable task synthesis with efficient RL~\citep{evocua15,uimate,scalecua_rl}. UI-TARS-2, Qwen-UI-Agent, and UI-Venus-2 go further and train one policy for mobile, computer, and web~\citep{uitars2,qwenuiagent,uivenus2}; to combine domains, a common recipe trains a separate RL expert per domain and merges the experts afterward~\citep{uitars2}. These works contribute models and recipes, and most build their own environments and training stack for the domains they study. \sys{} is complementary: it offers one open platform on which such recipes run across domains, and it trains mobile and computer tasks together in a single RL run.

\paragraph{GUI Gyms and Training Infrastructure.}
Open GUI gyms focus either on running agents or on supplying tasks (Table~\ref{tab:gym_overview}). Infrastructure-focused gyms make evaluation or training reproducible on one platform: BrowserGym unifies web benchmarks behind one interface, WebGym, MobileGym, and SCALECUA run online RL on hundreds of parallel environments, and ClawGUI provides a framework for training, evaluating, and deploying GUI agents~\citep{browsergym,webgym,mobilegym,scalecua_rl,clawgui}. Data-focused gyms, such as CUA-Gym, Gym-Anything, GUI-Genesis, and PhoneWorld, contribute new environments and tasks but release neither training code nor a way to run existing benchmarks~\citep{cuagym,gymanything,guigenesis,phoneworld}. Outside GUIs, AgentGym and GEM provide gyms for LLM agents in text- and tool-based environments~\citep{agentgym,gem}, and general-purpose RL frameworks such as verl, OpenRLHF, AReaL, and slime scale policy optimization for language models but do not manage GUI environments~\citep{verl,openrlhf,areal,slime}. Consistency is a further concern: changing the harness alone can change a fixed model's score~\citep{structagent}, so separate evaluation and training code can measure a different agent from the one being trained. \sys{} covers both sides on mobile, computer, and web. It runs ten existing benchmarks and online RL with $20{,}000$ parallel environments through one shared harness, builds its trainer on slime, and adds its own synthesized apps and tasks.

\paragraph{Environment and Task Synthesis.}
GUI benchmarks ship only hundreds of hand-written tasks, far fewer than online RL consumes, so recent work synthesizes tasks, and each approach trades reward reliability against scope. One line scores success with a VLM judge: ZeroGUI generates tasks and rewards them with a VLM, AutoPlay explores an app before proposing tasks with a VLM verifier, and OS-Genesis synthesizes trajectories in reverse~\citep{zerogui,autoplay,osgenesis}. These methods scale easily, but false-positive rewards can destabilize RL. A second line grounds rewards in code, which is more reliable but narrower. GUI-Genesis turns apps into lightweight web environments with executable assertions, AutoWebWorld models environments as finite-state machines, InfiniteWeb builds verifiable web tasks, and AutoEval derives rewards from mobile UI sub-states~\citep{guigenesis,autowebworld,infiniteweb,autoeval}. At larger scale, CUA-Gym builds grounded task pools for computers, MobileGym and PhoneWorld build mock mobile apps, and OpenComputer synthesizes desktop tasks and initial states around application-specific verifiers~\citep{cuagym,mobilegym,phoneworld,opencomputer}. In most of these systems, a task starts from a single input file or from sample data shared by all tasks of an app (Table~\ref{tab:gym_overview}). Task generation can also follow the learner: WebRL generates new tasks from failed attempts, TRACE turns recurrent agent failures into capability-targeted environments, and GenEnv and Agent-World evolve environments together with the agent~\citep{webrl,trace,genenv,agentworld}. \sys{} brings these threads together for GUI tasks. It builds a cross-app world of interlinked files and app records before writing any task, steers generation with diagnoses of failed rollouts, fixes success criteria before generating a code verifier, and accepts the verifier only after it passes execution tests. Its mocked apps include real-world clones, in which we measure transfer to the real app, and novel apps for studying adaptation to unfamiliar interfaces.

\section{Conclusion}
\label{sec:conclusion}

We presented \sys{}, an open all-in-one platform for GUI agent evaluation and online RL across mobile, computer, and web. Evaluation, data collection, and training share one rollout path and one agent harness, ensuring the evaluated agent is the trained agent; the environment manager serves ten benchmarks and has run $20{,}000$ parallel instances. \sys{} supplies online RL with synthesized mocked apps, cross-app worlds of interlinked files and app records, and tasks whose code verifiers pass execution tests. One joint RL run on mobile and computer tasks trained MAI-Gym-9B. It reaches $78.4\%$ on \mobileworld{} and $76.2\%$ on OSWorld-Verified, the best results among open GUI agents under 10B parameters; RL improves on the SFT warm-up across all five benchmarks outside its training environments. Our analyses indicate that joint training outperforms merging per-domain experts, larger synthesized task pools strengthen the policy, and skills learned in a real-world clone transfer to the real app, raising success on a physical phone from $28.8\%$ to $48.1\%$. We release the framework, synthesis skills, mocked apps, $2{,}000$ verified tasks, and the model checkpoint for GUI agent evaluation, synthesis, and training on one platform.

\section{Contributions}\label{contribution}

\subsection*{Core Contributors}
Panrong Tong$^{*}$, Quyu Kong$^{*}$, Pengxiang Li$^{*}$, Tianyu Xia$^{*}$, Hanzhang Zhou$^{*}$, Lei Wang, Chenglin Cai, Long Li, Long Chen, Jianan Zhang, Zhengwen Zeng, Xu Zhang, Gongjie Zhang, Yue Wang$^{\dagger}$, Steven Hoi.

\subsection*{Contributors}
Chen Liu, Yuchen Sun, Yang Zi, Peng Li, Nana Jiang, Jiaqi Gao, Jun Huang, Ennan Zhai, and Wenchao Zhou.

\begingroup
\renewcommand\thefootnote{}
\footnotetext{$^*$ Project Co-Leader. $^\dagger$ Project Leader.}
\endgroup

\clearpage
\bibliography{biblio}
\bibliographystyle{colm2024_conference_unsrt}

\clearpage
\appendix
\section{Appendix}

\subsection{The \sys{} Dashboard}
\label{app:dashboard}

\begin{figure}[!htbp]
\centering
\includegraphics[width=\linewidth]{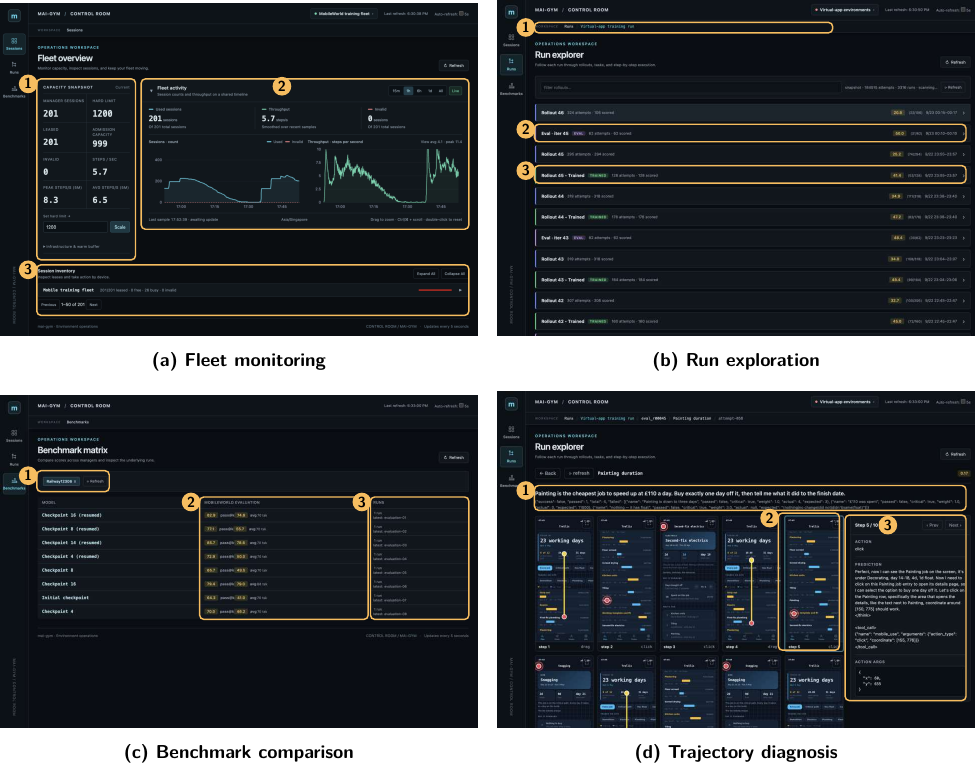}
\caption{\textbf{MAI-Gym dashboard.} \textbf{(a)}~Fleet capacity (1), usage
history (2), and sessions (3). \textbf{(b)}~Runs (1), evaluation rollouts (2),
and training batches (3). \textbf{(c)}~Benchmarks (1), scores (2), and runs (3).
\textbf{(d)}~Task and verifier result (1), action replay (2), and model output
(3).}
\label{fig:dashboard}
\end{figure}

Figure~\ref{fig:dashboard} shows fleet monitoring, run and benchmark navigation,
and step-by-step attempt replay.

\subsection{Gym Interface and Environment Contract}
\label{app:interface}
Every environment adapter implements five operations. \texttt{reqEnvs(spec, n)}
provisions instances; \texttt{observe(h)} returns the current observation;
\texttt{reset(h, x)} initializes an instance for task $x$; \texttt{step(h, a)}
applies an action; and \texttt{eval(h, x)} invokes the task verifier. The
environment manager implements allocation, session management, trajectory
recording, and reward delivery once above this contract;
Table~\ref{tab:contract} lists the responsibilities on each side.

\begin{table}[h]
\centering
\caption{\textbf{Responsibilities across the environment contract.}}
\label{tab:contract}
\footnotesize
\setlength{\tabcolsep}{4pt}
\begin{tabularx}{\linewidth}{@{}>{\raggedright\arraybackslash}p{0.15\linewidth}
                              >{\raggedright\arraybackslash}X
                              >{\raggedright\arraybackslash}X@{}}
\toprule
\textbf{Concern} & \textbf{Environment adapter} & \textbf{\sys{}} \\
\midrule
Capacity    & Reports health and available instances & Leases sessions, reclaims expired capacity, and assigns attempts \\
\addlinespace
Isolation   & Initializes task state and isolates concurrent instances & Keeps one attempt bound to one session for its lifetime \\
\addlinespace
Observation & Returns screenshots, command output, and metadata & Selects the transport and records observations in trajectories \\
\addlinespace
Action      & Executes the declared action vocabulary & Parses model output, dispatches actions, and records results \\
\addlinespace
Task        & Installs task state and invokes its verifier & Samples tasks, forms rollout groups, and delivers rewards to training \\
\bottomrule
\end{tabularx}
\end{table}

Sessions carry leases so that replaced workers cannot continue to act, and
mutating operations carry caller-supplied identifiers so that a retried request
does not apply the same action twice.

\subsection{Action Space}
\label{app:actions}
Table~\ref{tab:action-space} defines the action vocabulary used by \modelname{}.
It combines coordinate-based GUI operations with CLI and API calls, user
interaction, and task termination.

\begin{table}[t]
    \centering
    \caption{\textbf{Action space of \modelname{}.}}
    \label{tab:action-space}
    \small
    \setlength{\tabcolsep}{5pt}
    \begin{tabular}{@{}p{0.25\linewidth}p{0.67\linewidth}@{}}
        \toprule
        Action & Definition \\
        \midrule
        \multicolumn{2}{@{}l}{\textbf{GUI Actions}} \\
        \texttt{click} & Clicks at coordinates $(x,y)$. \\
        \texttt{double\_click} & Double-clicks at coordinates $(x,y)$; on touch screens, a double tap. \\
        \texttt{long\_press} & Long-presses at coordinates $(x,y)$. \\
        \texttt{type} & Types the specified text content. \\
        \texttt{hotkey} & Presses a key combination, such as \texttt{ctrl+c}. \\
        \texttt{open} & Opens the specified app.  \\
        \texttt{drag} & Drags from start coordinates $(x_1,y_1)$ to end coordinates $(x_2,y_2)$. \\
        \texttt{system\_button} & Presses a system button, selected from \texttt{back}, \texttt{home}, \texttt{menu}, and \texttt{enter}. \\
        \texttt{wait} & Waits for a specified duration in seconds. \\
        \midrule
        \multicolumn{2}{@{}l}{\textbf{CLI Actions}} \\
        \texttt{cli\_command} & Executes a bash command in the active CLI environment. \\
        \midrule
        \multicolumn{2}{@{}l}{\textbf{API Actions}} \\
        \texttt{api\_call} & Invokes an API with specified arguments. \\
        \midrule
        \multicolumn{2}{@{}l}{\textbf{Interaction and Control Actions}} \\
        \texttt{ask\_user} & Requests information from the user. \\
        \texttt{terminate} & Ends the task with a final answer and a status of \texttt{success} or \texttt{failed}. \\
        \bottomrule
    \end{tabular}
\end{table}

\subsection{Hyperparameters}
\label{app:hparams}
Unless stated otherwise, runs use the settings below; Section~\ref{sec:exp} notes
settings that differ for individual experiments, and Appendix~\ref{app:config}
gives the shared configuration structure.

\subparagraph{Initialization.}
The policy is a Qwen3.5-9B backbone and keeps that model's tokenizer.
Observations are screenshots at the environment's native resolution, together
with command output on computers; actions are emitted in the
vocabulary of Appendix~\ref{app:actions}. Before the RL stage the model is warmed
up by supervised fine-tuning on mixed mobile and computer data. The joint RL run
and the expert runs of Section~\ref{sec:ana:joint} start from this warm-up checkpoint.

\subparagraph{Optimization.}
Unless an experiment says otherwise, the algorithm is GRPO with a group size of
$8$ and a per-token loss. Every policy-generated token of trajectory $i$ receives
the advantage $A_i$ of Equation~\ref{eq:grpo-adv}. With $\mathcal{Y}_i$ the
policy-generated tokens of trajectory $i$ and
$\rho_u=\pi_\theta(u\mid\cdot)/\pi_{\theta_{\mathrm{old}}}(u\mid\cdot)$ the
importance ratio of token $u$, the policy maximizes the clipped
surrogate~\citep{ppo}
\begin{equation}
\label{eq:grpo-obj}
    \mathcal{L}(\theta)=\frac{1}{\sum_{i}|\mathcal{Y}_i|}\sum_{i=1}^{G}\sum_{u\in\mathcal{Y}_i}
    \min\!\big(\rho_u A_i,\ \operatorname{clip}(\rho_u,1-\epsilon,1+\epsilon)\,A_i\big).
\end{equation}
The objective carries a KL loss term with coefficient
$10^{-3}$ and no entropy bonus; the reward itself is not KL-penalized, so the
constraint acts on the loss rather than on the returns. Adam runs at a constant
learning rate of $10^{-6}$, with weight decay $0.1$ and
$(\beta_1, \beta_2) = (0.9, 0.98)$. The micro-batch is a single sequence, and the
unit the trainer counts is the window, not the step or the trajectory, which
follows from the window-wise packing of Section~\ref{sec:training}. History uses
a window of $K=10$ steps with an overlap of $O=2$. Training runs on Megatron-LM
and generation on SGLang, with the parallelism of Appendix~\ref{app:config}.

\subparagraph{Task Selection.}
Within a pool, sampling follows the policy-adaptive rule of
Section~\ref{sec:training}: a success threshold of $\eta=0.8$ separates mastered
tasks from the frontier, and never-solved and mastered tasks receive reduced
budgets of $4$ and $2$ attempts against the full group of $8$. In the joint run, a
rollout round draws from both domains; the pools themselves are the output of the synthesis pipeline
of Section~\ref{sec:taskenv}, and Appendix~\ref{app:taskexamples} shows what its
tasks look like.

\subparagraph{Evaluation.}
Every benchmark is run through the harness of Section~\ref{sec:observability}
with that benchmark's own verifier and step limit (Table~\ref{tab:benchmarks}).
Decoding uses temperature 0.0 with 3 trials per task.

\subsection{Training Configuration}
\label{app:config}
A single YAML file configures a training run (Section~\ref{sec:training}): the task pools and the environments behind them, the algorithm, the rollout budget, the history format, and the training backend. The listing below follows the structure of the file the runs in this paper use. Values in angle brackets are set per experiment; credentials are never written in the file and are read from the environment at load time.

\definecolor{cfgbg}{RGB}{244,247,245}
\definecolor{cfgaccent}{RGB}{61,122,106}
\definecolor{cfgkey}{RGB}{47,110,94}
\definecolor{cfgmuted}{RGB}{128,136,133}
\lstdefinestyle{gymyaml}{basicstyle=\def\ttdefault{zi4}\ttfamily\footnotesize,
  columns=fixed, basewidth=0.5em, keepspaces=true, showstringspaces=false,
  aboveskip=0pt, belowskip=0pt,
  comment=[l]{\#}, commentstyle=\color{cfgmuted},
  moredelim=[s][\color{cfgmuted}]{<}{>},
  alsoletter={_}, morekeywords={model,checkpoint,policy_variant,tool_parser,tasks,pool,env_type,algorithm,name,group_size,learning_rate,reward_allocation,sampling,success_threshold,fail_budget,mastered_budget,rollout,concurrency,num_sessions,temperature,max_turns,max_total_length,history,window,overlap,image_retention,token_native_history,windowwise_train,observations,screenshot_transport,oss,endpoint,region,access_key_id,access_key_secret,trainer,backend,parallelism,tensor,pipeline,context,data,kl_loss_coef,entropy_coef,per_token_loss},
  keywordstyle=\color{cfgkey}}
\begin{tcblisting}{listing only, enhanced, breakable,
    colback=cfgbg, frame hidden, arc=2pt,
    borderline west={2.2pt}{0pt}{cfgaccent},
    left=10pt, right=6pt, top=4pt, bottom=4pt,
    listing options={style=gymyaml}}
model:
  checkpoint: <SFT warm-up checkpoint>
  policy_variant: <action/response format>
  tool_parser: <action parser>

tasks:                           # pools, and the environment each is served by
  - pool: <mobile task pool>
    env_type: <mobile backend>
  - pool: <computer task pool>
    env_type: <computer backend>

algorithm:
  name: grpo                     # or sao, ...
  group_size: 8
  learning_rate: 1.0e-6
  kl_loss_coef: 1.0e-3           # KL on the loss, not on the reward
  entropy_coef: 0.0
  per_token_loss: true
  reward_allocation: uniform     # terminal advantage on every policy token

sampling:                        # policy-adaptive task sampling
  success_threshold: 0.8         # eta
  fail_budget: 4                 # attempts for never-solved tasks
  mastered_budget: 2             # attempts for mastered tasks

rollout:
  concurrency: <C>               # trajectories in flight
  num_sessions: <leased sessions per pool>
  temperature: <temperature>
  max_turns: <step limit>
  max_total_length: <context budget per sample>

history:
  window: 10                     # K
  overlap: 2                     # O
  image_retention: <screenshots kept>
  token_native_history: true     # replay the sampled tokens verbatim
  windowwise_train: true         # each window becomes one training sample

observations:
  screenshot_transport: oss      # inline, or by reference to object storage

trainer:
  backend: <training backend>
  parallelism: {tensor: <tp>, pipeline: <pp>, context: <cp>, data: <dp>}
\end{tcblisting}

\FloatBarrier

\subsection{Synthesized Task Examples}
\label{app:taskexamples}

\begin{figure}[!hb]
\centering
\definecolor{ebA}{RGB}{47,102,122}
\setlength{\fboxsep}{3pt}\setlength{\fboxrule}{0.4pt}
\begin{minipage}[t]{0.485\linewidth}
\vspace{0pt}
{\small\color{ebA}\bfseries (a) Comparing print proofs}\par\smallskip
\fcolorbox{ebA!30}{white}{\begin{minipage}[c][4.45cm][c]{\dimexpr\linewidth-6.8pt\relax}\centering
\includegraphics[width=\linewidth,height=4.40cm,keepaspectratio]{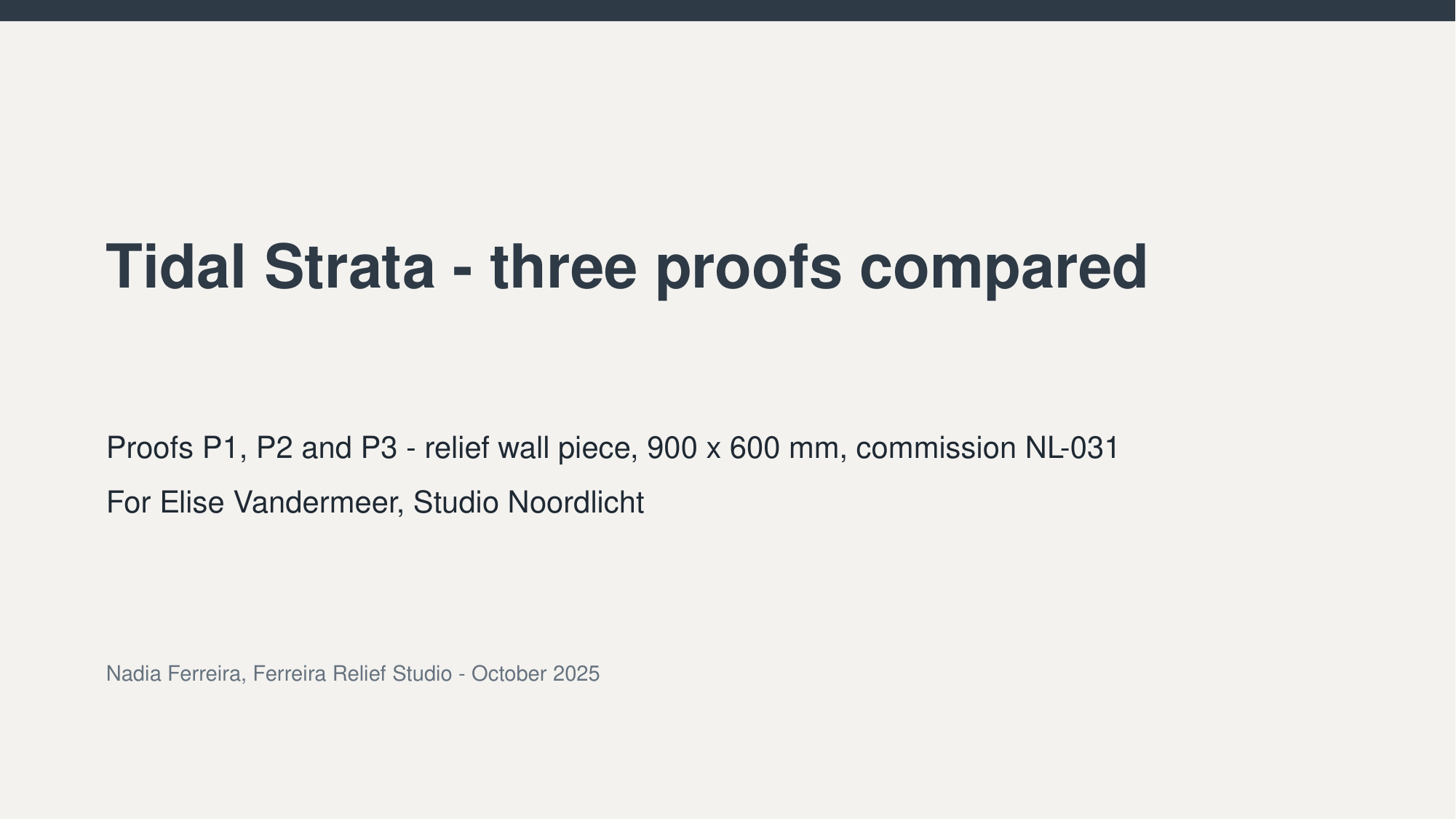}
\end{minipage}}
\par\smallskip
{\scriptsize\itshape Initial title slide}\par\medskip
\footnotesize\raggedright
\textbf{Task.} Read the job ticket, handwritten notes, and confirmation email to
consolidate the printing settings for three proofs in Calc. Add one slide per
proof to the comparison deck. Export the spreadsheet to CSV and the completed
presentation to PDF.

\medskip
\textbf{Input Files.} Draft presentation, job ticket, studio notes, and the print
lab's confirmation email.

\medskip
\textbf{Checks.} CSV and presentation values match the required settings.
Each proof has its own slide, and the PDF includes all three proof labels.
\end{minipage}\hfill
\begin{minipage}[t]{0.485\linewidth}
\vspace{0pt}
{\small\color{ebA}\bfseries (b) Completing a symposium deck}\par\smallskip
\fcolorbox{ebA!30}{white}{\begin{minipage}[c][4.45cm][c]{\dimexpr\linewidth-6.8pt\relax}\centering
\includegraphics[width=\linewidth,height=4.40cm,keepaspectratio]{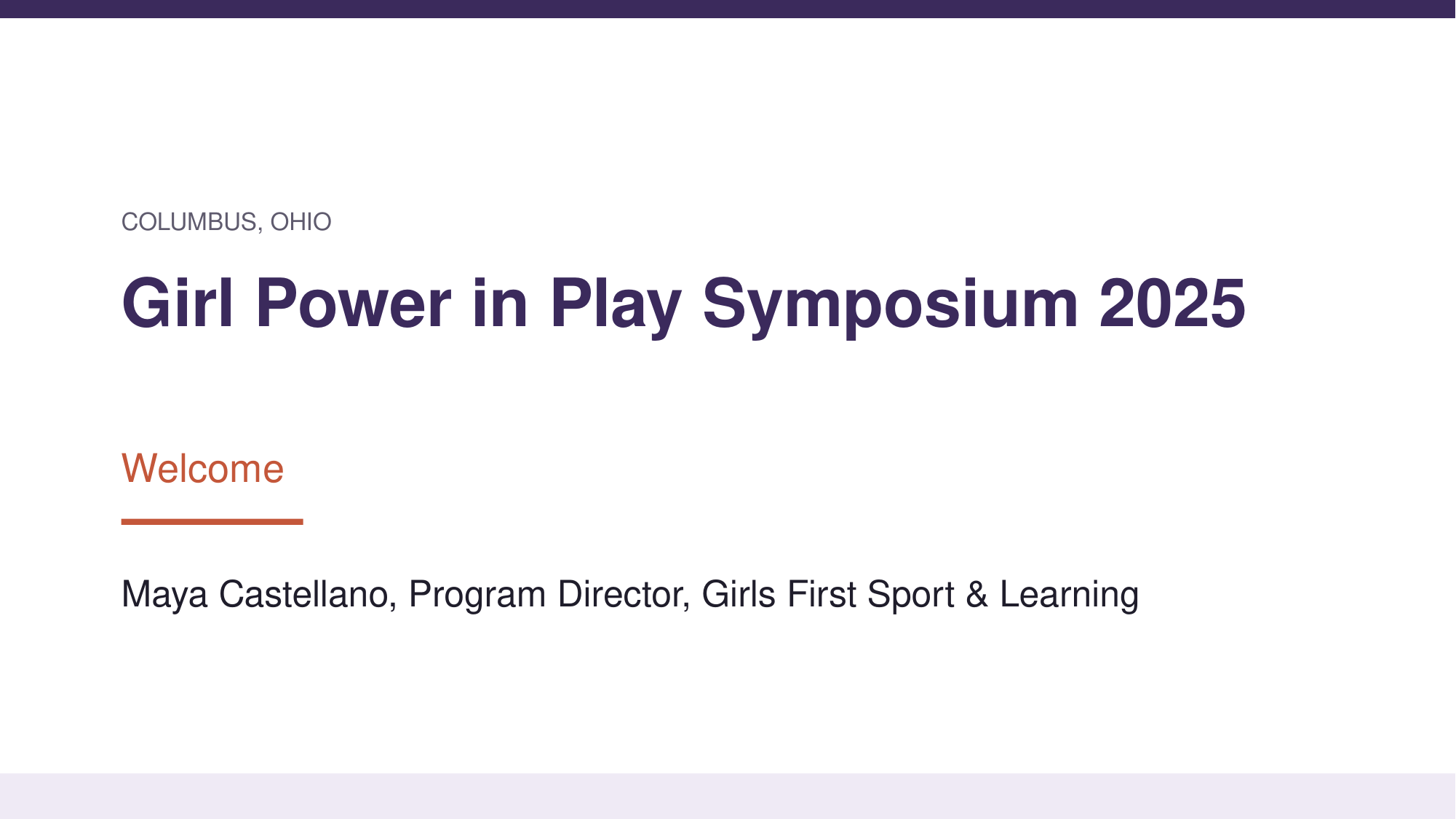}
\end{minipage}}
\par\smallskip
{\scriptsize\itshape Initial title slide}\par\medskip
\footnotesize\raggedright
\textbf{Task.} Complete the symposium opening deck using the planning files.
Add the proposed date and venue, each speaker's role and home city, and the
budget items. Build a budget bar chart in Calc and include it in the deck.
Compute total speaker travel cost from fares and hotel costs, and save the
completed presentation as a new file.

\medskip
\textbf{Input Files.} Draft presentation, planning documents, budget workbook,
and speaker travel estimates.

\medskip
\textbf{Checks.} Required slides contain the venue, speaker, and budget facts;
the embedded chart has the correct categories and values; the summary reports
the computed travel total.
\end{minipage}
\caption{\textbf{Synthesized desktop task examples.} Screenshots show the
initial slides; task instructions and verifier checks are summarized.}
\label{fig:taskexamples}
\end{figure}

Figure~\ref{fig:taskexamples} shows two presentation tasks from synthesized
desktop workspaces. Comparing print proofs requires settings from a job ticket,
handwritten notes, and a confirmation email. Completing the
symposium deck uses planning documents, a budget workbook,
and speaker travel estimates. The verifiers check slide text,
dates, chart data, and computed totals in the saved presentations.

\begin{figure}[!t]
\centering
\definecolor{ebB}{RGB}{61,122,106}
\definecolor{ebBbg}{RGB}{244,249,246}
\setlength{\fboxsep}{5pt}\setlength{\fboxrule}{0.4pt}
\begin{minipage}[t]{0.46\linewidth}
\vspace{0pt}
\fcolorbox{ebB}{ebBbg}{\begin{minipage}[t]{\dimexpr\linewidth-10.8pt\relax}\footnotesize\raggedright
{\color{ebB}\bfseries Mobile: street-tree inventory}\par\smallskip
Ramona Delgado wants our street tree inventory attached to the Kestrel Ridge
application, topped with a one-page summary. Analyze the three surveys in your
Downloads folder [\ldots] using the methodology in
Street\_Tree\_Inventory\_Protocol\_2023.pdf, and save the summary as a new
spreadsheet [\ldots] one row per street plus a totals row, containing the
figures requested in her email. Finally, reply to her email [\ldots] and provide
each street's figures, the totals, and the date each street was last surveyed.
\end{minipage}}
\end{minipage}\hfill
\begin{minipage}[t]{0.50\linewidth}
\vspace{0pt}
{\scriptsize\color{ebB}\bfseries The protocol named in the instruction}\par\smallskip
\fcolorbox{ebB!40}{white}{\includegraphics[width=\dimexpr\linewidth-10.8pt\relax]{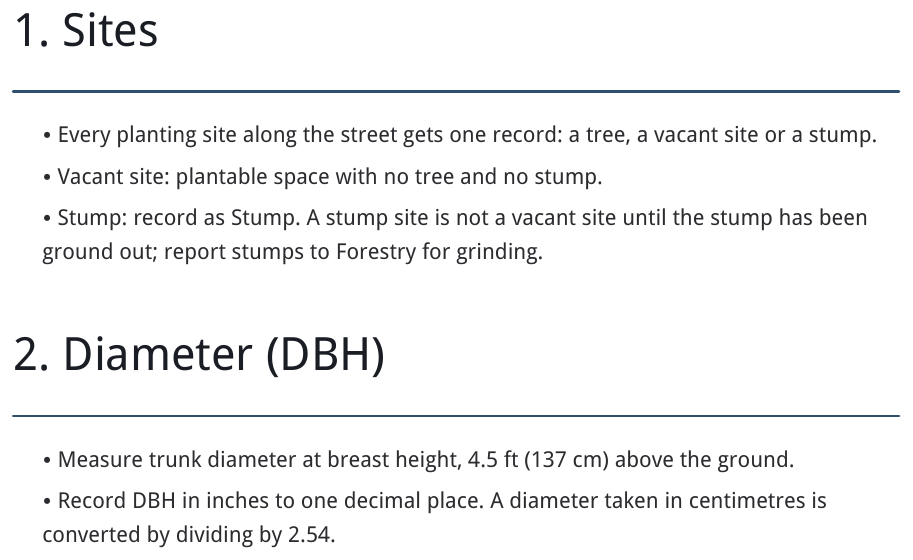}}
\end{minipage}
\caption{\textbf{A mobile task with a generated protocol.} Left: abridged
street-tree inventory instruction. Right: the protocol specifying counting and
measurement rules.}
\label{fig:taskexamplesmobile}
\end{figure}

Figure~\ref{fig:taskexamplesmobile} shows a mobile example: the street-tree
inventory task applies counting and unit-conversion rules from a generated
survey protocol. In each case, task files determine the required result.

\subsection{Mocked Apps}
\label{app:mockapps}

We release seven real-world clones and $42$ novel apps.

\subsubsection{Real-World Clones}
\label{app:mockedapps}

We build resettable clones of seven apps whose real versions move money, place
orders, or publish posts. Each clone reproduces the main user flows over a
backend that restores its state between attempts.

Testers carried out the same tasks in each clone and its real counterpart and
compared the screens along the way. Figure~\ref{fig:mocked-apps} in
Section~\ref{sec:taskenv:env} shows these comparisons. The China Construction
Bank tasks require a real bank card, so its pair shows the home screens.

\subsubsection{Novel Apps}
\label{app:virtualapps}

The $42$ novel apps contain $840$ tasks at three difficulty levels. Each app is
a deterministic state machine executed in Chromium, and each task specifies an
instruction, initial state, step limit, executable completion checks, and a
reference action sequence. The collection contains $3{,}196$ checks over app
state, action history, and submitted answers.

Before release, every reference sequence must complete its task within the step
limit. The validation suite also checks reset restoration, rejects untouched and
random-action states, and verifies agreement between browser and browser-free
execution.

\end{document}